\documentclass[runningheads]{llncs}

\usepackage{eccv}

\usepackage{eccvabbrv}

\usepackage{amsmath,amsfonts,bm}

\def\Figref#1{Figure~\ref{#1}}

\def\Secref#1{Section~\ref{#1}}

\def\eqref#1{equation~\ref{#1}}

\def\1{\bm{1}}

\DeclareMathAlphabet{\mathsfit}{\encodingdefault}{\sfdefault}{m}{sl}
\SetMathAlphabet{\mathsfit}{bold}{\encodingdefault}{\sfdefault}{bx}{n}

\usepackage{amssymb}

\usepackage{booktabs}
\usepackage{colortbl}
\usepackage{multirow}
\usepackage{multicol}
\usepackage{makecell}

\usepackage{graphicx}
\usepackage{subcaption}
\usepackage{tikz}

\usepackage{microtype}
\usepackage{xcolor}
\usepackage{pifont}
\usepackage{enumitem}

\usepackage[accsupp]{axessibility}

\usepackage{hyperref}

\usepackage{orcidlink}



\newcommand{\frameworkname}{Affogato\xspace}
\newcommand{\dataname}{Affogato-750K\xspace}
\newcommand{\modelname}{Espresso\xspace}

\newcommand{\Tref}[1]{Table~\ref{#1}}

\newcommand{\noindentbold}[1]{\noindent \textbf{#1.}\xspace}

\definecolor{darkyellow}{rgb}{0.92,0.75,0.0}

\DeclareRobustCommand{\ccircle}[1]{\tikz[baseline=-0.6ex]{\draw[black,fill=#1] (0,0) circle (0.7ex);}}

\begin{document}

\title{Affogato: Open-Vocabulary Affordance Grounding with Automated Data Generation at Scale}

\titlerunning{Affogato: Open-Vocabulary Affordance Grounding}

\author{Junha Lee\inst{1,2}$^{*}$\orcidlink{0009-0002-0217-8294} \and
Eunha Park\inst{1}$^{*}$\orcidlink{0009-0005-6062-0441} \and
Chunghyun Park\inst{1}\orcidlink{0000-0001-9743-8495} \and \\
Dahyun Kang\inst{1}\orcidlink{0000-0001-9252-9582} \and
Minsu Cho\inst{1,3}\orcidlink{0000-0001-7030-1958}}

\authorrunning{J.~Lee et al.}

\institute{Pohang University of Science and Engineering (POSTECH)  \and SqueezeBits \and RLWLRD\\
\url{https://junha-l.github.io/affogato/}}

\maketitle
\let\thefootnote\relax\footnotetext{$^{*}$Equal contribution.}

\begin{abstract}

Affordance grounding aims to localize where to interact with an object, a fundamental capability for embodied agents. Yet progress is bottlenecked by data: manual annotation is prohibitively expensive and confines existing datasets to a narrow set of predefined object and affordance categories.
We introduce \textbf{\frameworkname}, a framework for open-vocabulary affordance grounding centered on \textbf{\dataname}, a large-scale dataset of 750K 3D affordance heatmaps paired with natural language queries.
We build it with a fully automated pipeline that orchestrates foundation models to generate them at scale without human labeling.
It covers significantly more diverse categories than any existing dataset. For reliable evaluation, we further provide 5K human-verified test pairs.
We also present \textbf{\modelname-3D} and \textbf{\modelname-2D}, simple yet effective models with a unified architecture across both modalities.
Pretraining on \dataname improves both \modelname and prior methods and yields the largest gains on unseen object and affordance categories, showing that it provides broadly transferable supervision across architectures.

\end{abstract}

\section{Introduction}

\begin{figure}[t]
    \centering
    \includegraphics[width=\linewidth]{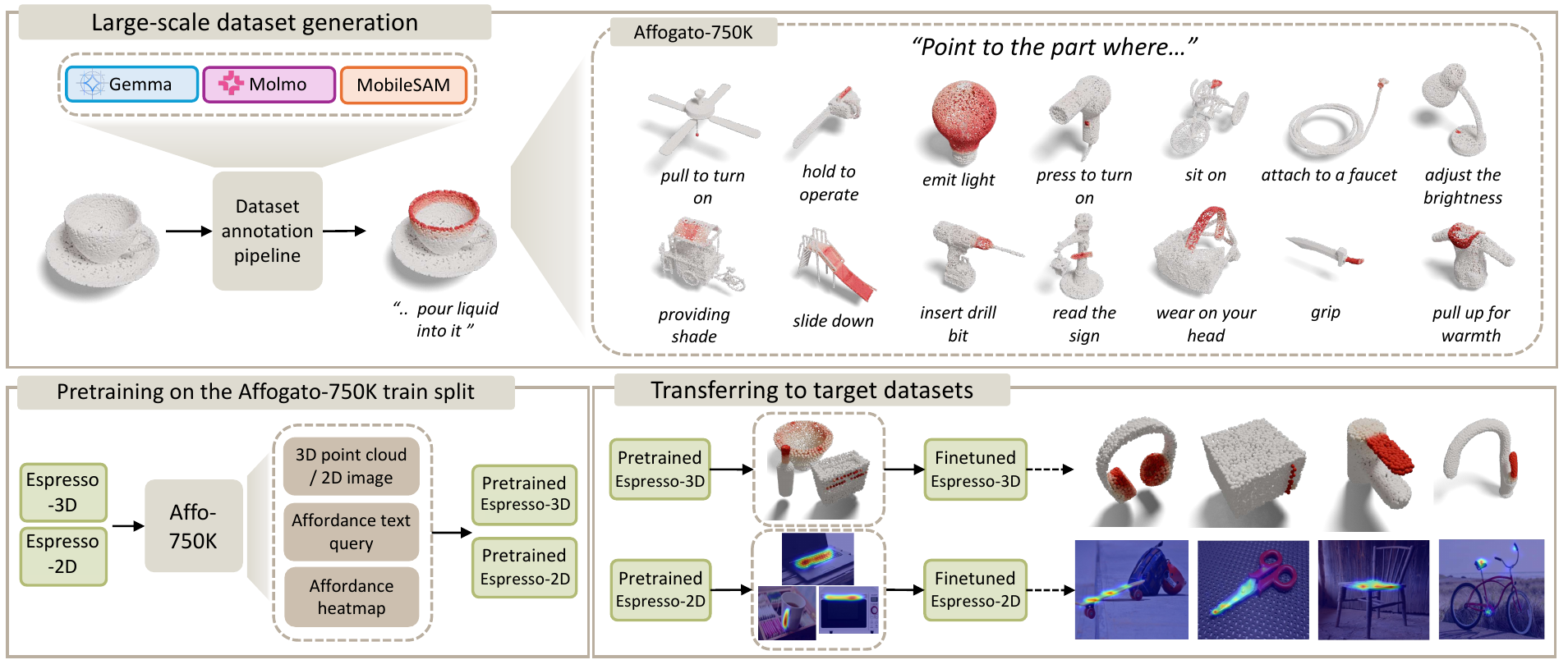}
    \caption{\textbf{Overview of \frameworkname.} At the heart of \frameworkname is \textbf{\dataname} (dataset), validated through \textbf{\modelname} (models). \dataname comprises 750K open-vocabulary affordance annotations across 150K 3D objects, built via an automated pipeline chaining Gemma3~\cite{team2025gemma} (query generation), Molmo~\cite{deitke2025molmo} (point prediction), and MobileSAM~\cite{mobile_sam} (mask generation) with multi-view voting for 3D aggregation. The test split is human-verified to ensure evaluation quality. \modelname-3D and \modelname-2D follow a unified architecture and demonstrate substantial pretraining gains for \modelname models and existing methods.}
    \label{fig:teaser}
\end{figure}

The theory of affordances addresses what an environment offers, provides, or furnishes animals~\cite{gibson2014ecological}. 
For example, what does a cup afford to humans?
The answer can be drinking, lifting, throwing, \etc.
While the concept of affordances has long been studied in psychology~\cite{asada2009cognitive, jamone2016affordances} and robotics~\cite{chemero2007gibsonian, min2016affordance,ardon2020affordances}, here we focus on \textit{visual affordance grounding} from a computer vision perspective~\cite{hassanin2021visual}; 
given an object as a 3D point cloud or 2D image, the task is to localize the region where a described interaction takes place.
Unlike other types of visual grounding/segmentation~\cite{yu2016modeling,kang2024defense,lisa}, affordance grounding is particularly challenging for two main reasons.
First, an object may afford multiple functionalities across different regions, and these regions often have indistinct boundaries, making localization inherently ambiguous. Second, affordances are open-vocabulary by nature, and a fixed set of predefined terms cannot describe them adequately.

Addressing these challenges requires affordance supervision that is open-vocabulary, captures the soft and ambiguous extent of affordance regions, and can be generated at scale without exhaustive manual annotation.
However, existing datasets fail to capture this open-ended nature; current 3D datasets~\cite{deng20213d,nguyen2023open,yang2023grounding,li2024laso,chu20253d,yu2025seqafford} are constrained to a small set of predefined object and affordance categories, and 2D datasets~\cite{myers2015affordance,nguyen2017object,chuang2018learning,fang2018demo2vec,nagarajan2019grounded,agd20k,mur2023multi} remain similarly limited in coverage (\Tref{tab:dataset_comparison}).
This limitation stems from their reliance on manual annotation, which cannot scale to the diversity of real-world objects and interactions.

To overcome these limitations, we introduce \textbf{\frameworkname} (\textit{AFFOrdance Grounding All aT Once}), a scalable framework for open-vocabulary affordance grounding (Figure~\ref{fig:teaser}).
At its core is \textbf{\dataname}, a large-scale dataset that directly addresses the scarcity of open-vocabulary affordance data.
It is built fully automatically with an annotation pipeline that orchestrates the complementary strengths of multiple foundation models~\cite{team2025gemma,deitke2025molmo,mobile_sam}. The pipeline generates open-vocabulary affordance queries, localizes and segments the corresponding interaction regions in each view, and aggregates these per-view predictions into robust 3D heatmaps.

\dataname comprises 150K 3D objects from Objaverse~\cite{objaverse}, each annotated with five affordance queries and corresponding 3D heatmaps, totaling 750K annotations. We further derive a 2D split by projecting the heatmaps onto rendered views, and curate a test split of 5K human-verified pairs for reliable evaluation. As shown in \Tref{tab:dataset_comparison}, \dataname significantly surpasses existing datasets in both scale and category diversity.

Alongside the dataset, we design \modelname-3D and \modelname-2D, minimal and symmetric models built from a vision encoder, a text encoder, and a heatmap decoder. Despite this simplicity, \modelname achieves strong affordance grounding performance in both modalities. Experiments demonstrate that pretraining on \dataname yields substantial and consistent gains for both \modelname and existing methods, confirming the dataset's effectiveness as a scalable supervision source across modalities.

To summarize, our contributions are as follows:
\begin{table}[!t]
    \caption{Comparison of 3D (left) and 2D (right) affordance grounding datasets. Aff abbreviates affordances. For \dataname, classes and affordances are Gemma3~\cite{team2025gemma} predictions with frequencies above 10 and 100, respectively.}
    \label{tab:dataset_comparison}
    \begin{minipage}[t]{.56\linewidth}
      \centering
    \resizebox{\linewidth}{!}{
        \begin{tabular}{lrrrr}
            \toprule
            Dataset & \# Classes & \# Aff & \# 3D assets & \# Questions \\
            \midrule
            3D AffordanceNet~\cite{deng20213d} & 23 & 18 & 22,949 & 0 \\
            PartAfford~\cite{xu2022partafford} & 23 & 24 & >25,000 & 0 \\
            OpenAD~\cite{nguyen2023open} & 23 & 37 & 22,949 & 0 \\
            PIAD~\cite{yang2023grounding} & 23 & 17 & 7,012 & 0 \\
            LASO~\cite{li2024laso} & 23 & 17 & 8,434 & 870 \\
            SceneFun3D~\cite{delitzas2024scenefun3d} & N/A & ~~9 & 710 & 17,133 \\
            IRAS~\cite{chu20253d} & 23 & 36 & 22,949 & 42,119 \\
            SeqAfford~\cite{yu2025seqafford} & 23 & 18 & 18,371 & 162,386 \\
            \rowcolor{gray!15} \dataname & \cellcolor{gray!15}\textbf{>450} & \cellcolor{gray!15}\textbf{>350} & \textbf{150,104} & \textbf{750,520} \\
            \bottomrule
        \end{tabular}
    }
    \end{minipage}%
    \hfill
    \begin{minipage}[t]{.44\linewidth}
          \centering
    \resizebox{\linewidth}{!}{
        \begin{tabular}{lrrrrr}
            \toprule
            Dataset  & \# Classes & \# Aff & \# 2D images & \# Questions\\
            \midrule
            UMD~\cite{myers2015affordance} & 17 & 7 & 10,000 & 0 \\
            IIT-Aff~\cite{nguyen2017object} & 10 & 9 & 8,835 & 0 \\
            ADE-Aff~\cite{chuang2018learning} & 150 & 3 & 10,011 & 0 \\
            ORPA~\cite{fang2018demo2vec} & N/A & 7 & 2,512 & 0 \\
            Grounded I.H.~\cite{nagarajan2019grounded} & 31 & 20 & 1,871 & 0 \\
            AGD20K~\cite{agd20k} & 50 & 36 & 23,816 & 0 \\
            EPIC-Aff~\cite{mur2023multi} & 304 & 20-36 & 38,876 & 0 \\
            3DOI~\cite{qian2023understanding} & N/A & 3 & 10,000 & 0 \\
            VRB~\cite{bahl2023affordances} & N/A & N/A & 54,000 & 0 \\
            2HandedAfforder~\cite{heidinger20252handedafforder} & 163 & 73 & \textbf{278,000} & 278,000 \\
            RAGNet~\cite{wu2025ragnet} & 180 & N/A & 273,000 & 26,000 \\
            \rowcolor{gray!15} \dataname & \cellcolor{gray!15}\textbf{>450} & \cellcolor{gray!15}\textbf{>350} &
            150,104 & \textbf{750,520} \\
            \bottomrule
        \end{tabular}
    }
    \end{minipage}
\end{table}

\begin{itemize} [leftmargin=1.5em]
    \item We introduce \textbf{\frameworkname}, a scalable framework for open-vocabulary affordance grounding, built on \textbf{\dataname}, a large-scale dataset of 750K affordance annotations over 150K 3D instances that covers far more diverse object and affordance categories than existing datasets, with a 5K human-verified test split for reliable evaluation.
    \item We develop a fully automated pipeline that orchestrates foundation models to generate reliable query-heatmap pairs at scale without manual labeling.
    \item We present \textbf{\modelname-3D} and \textbf{\modelname-2D}, simple yet effective models with a unified design across 3D and 2D, and show that pretraining on \dataname yields substantial gains for both \modelname and prior methods.
\end{itemize}

\section{Related work}
\noindentbold{Visual affordance grounding}
Visual affordance grounding aims to localize functional regions in visual data given affordance concepts.
Early 2D approaches~\cite{myers2015affordance, nguyen2017object, chuang2018learning, fang2018demo2vec, nagarajan2019grounded, agd20k, mur2023multi, qian2023understanding, bahl2023affordances} focused on predefined affordance categories with limited object diversity.
To overcome the scarcity of annotations, recent works explored weakly supervised learning~\cite{agd20k, tang2025closed} and few-shot or zero-shot approaches~\cite{li2024one, cuttano2024does}. 
Other works leveraged external videos of human–object interactions~\cite{fang2018demo2vec, nagarajan2019grounded, liu2022joint, mur2023multi, ju2024robo, heidinger20252handedafforder} to generate affordance heatmaps. 
However, supervision derived from such videos is often noisy and biased due to occlusions and limited viewpoints.
Although recent advances in automation have enabled the construction of large-scale 2D datasets~\cite{heidinger20252handedafforder, wu2025ragnet}, annotations based solely on single-view observations are inherently constrained compared to annotations aggregated from multi-view evidence.

In the 3D domain, 3D AffordanceNet~\cite{deng20213d} established the first closed-vocabulary 3D affordance grounding benchmark using PartNet~\cite{mo2019partnet} shapes. PartAfford~\cite{xu2022partafford} and OpenAD~\cite{nguyen2023open} expanded affordance vocabularies while maintaining categorical constraints. Recent advances toward open-vocabulary 3D grounding include LASO~\cite{li2024laso}, which introduced free-form textual queries with 3D heatmap prediction, and 3D AffordanceLLM~\cite{chu20253d}, which leveraged language models for query generation. SeqAfford~\cite{yu2025seqafford} focused on textual diversity through GPT-4 generated questions, and LMAffordance3D~\cite{zhu2025lmaffordance3d} fused 2D and 3D features with language guidance. At the scene level, SceneFun3D~\cite{delitzas2024scenefun3d} explored functionality segmentation, and Fun3DU~\cite{corsetti2025functionality} proposed a training-free framework leveraging VLMs for 3D scene affordance. More recently, methods that transfer 2D semantic knowledge to 3D affordance segmentation have also been explored~\cite{huang2026unlocking,he2026task}.

Despite progress, existing datasets remain severely constrained in scale and category diversity (\Tref{tab:dataset_comparison}). 3D datasets share the same 23 object classes derived from a single source~\cite{mo2019partnet}, while 2D datasets cover at most 304 object and 73 affordance categories. All existing datasets rely on predefined closed-set labels, fundamentally limiting generalization to open-vocabulary queries. This data scarcity is the primary bottleneck for affordance grounding generalization.

\noindentbold{Harnessing 2D foundation models for 3D supervision}
Recent advances in 2D foundation models have opened new pathways for generating high-quality 3D supervision without manual annotation. 
Several works~\cite{luo20223d,xue2024ulip,xu2024pointllm} leverage large language models (LLMs) and vision-language models (VLMs) to generate text captions for 3D objects using their multi-view rendered images, facilitating joint 3D-language learning. 
Others~\cite{yang2024sampart3d,liu2025partfield} distill 2D visual knowledge from foundation models to train 3D geometric encoders,
while scene-centric methods~\cite{yang2024regionplc,weder2024labelmaker,lee2025mosaic3d} extend this paradigm to large-scale environments by generating region-level 3D annotations through integration with 2D foundation models.
Our work extends this paradigm to functional affordance understanding by leveraging 2D foundation models to generate 3D affordance supervision. A vision-language model generates affordance queries and interaction points, MobileSAM~\cite{mobile_sam} produces 2D segmentation masks, and the resulting masks are lifted to 3D and aggregated via multi-view voting to obtain reliable 3D affordance heatmaps.

\section{The \dataname dataset}

Advancing human-object interaction understanding in embodied AI systems demands comprehensive affordance grounding data—a critical resource currently lacking in the field. 
Existing datasets~\cite{li2024laso,deng20213d} suffer from key limitations in scale, diversity, and annotation quality, creating a substantial barrier to progress in this domain.
To address the issue, we propose the \textit{\dataname} dataset, a large-scale open-vocabulary affordance grounding dataset. 
In this section, we elaborate on the pipeline used to generate \dataname.
Our pipeline leverages large-scale 3D object repositories~\cite{objaverse} and state-of-the-art foundation models~\cite{mobile_sam,team2025gemma,deitke2025molmo} to automatically generate high-quality affordance annotations.
This addresses the aforementioned limitations by enabling robust and generalizable affordance learning across diverse objects.

\subsection{Source dataset}
\label{subsec:source}
We build the \dataname dataset upon Objaverse~\cite{objaverse}, one of the largest public 3D asset repositories. It contains more than 700K web-crawled 3D object meshes spanning diverse object categories and geometries.
Since recent vision-language models (VLMs) take 2D images and text as input and cannot directly process 3D meshes, we use multi-view renderings of each 3D object as an intermediary representation that bridges 3D geometry and language.
To this end, we utilize G-Objaverse~\cite{zuo2024sparse3d}, which already provides high-resolution multi-view renderings, along with depth maps and camera parameters, for over 280K Objaverse objects.
Specifically, we select four subsets that have strong relevance to human-object interaction and functional affordances in daily lives: \textit{Daily-Used}, \textit{Furnitures}, \textit{Transportations}, and \textit{Electronics}, which results in 150K 3D objects.

\subsection{Annotation pipeline}
\label{subsec:data_pipeline}
Our data annotation pipeline consists of three stages. Given a textured 3D object as input, it outputs a set of natural language affordance queries alongside spatially localized 3D affordance heatmaps. 
The entire process is automated and designed to scale to hundreds of thousands of objects, making it suitable for constructing large-scale datasets. 
The overall annotation process is illustrated in \Figref{fig:data_pipeline}. 
Stage-wise qualitative results are provided in the supplementary material.
\begin{figure}[t!]
    \centering
    \includegraphics[width=\linewidth]{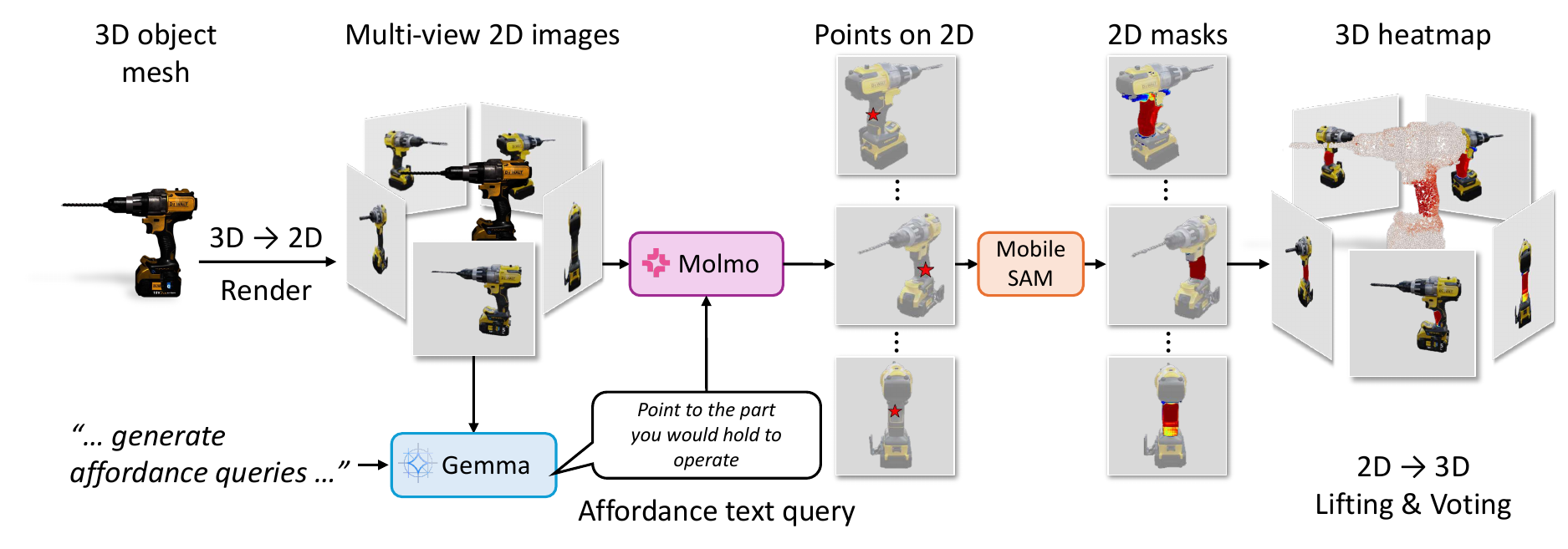}
    \caption{\textbf{Overview of our data annotation pipeline.}
    Given multi-view renderings of an object, Gemma3~\cite{team2025gemma} generates affordance queries, Molmo~\cite{deitke2025molmo} points the affordance, and MobileSAM~\cite{mobile_sam} converts the point into an affordance heatmap. The multi-view 2D affordance heatmaps are then aggregated on the 3D object surface to obtain a final 3D affordance heatmap.
    }
    \label{fig:data_pipeline}
\end{figure}

\noindentbold{Stage 1: Open-vocabulary affordance query generation}
Given multi-view images of a 3D object, we employ a vision-language model, Gemma3~\cite{team2025gemma}, to produce a set of natural language queries (five per object) that describe how a human might interact with the object.
These queries follow a constrained yet expressive format, allowing open-vocabulary interaction descriptions while maintaining spatial grounding. 
By conditioning on rendered views instead of object class labels~\cite{li2024laso,chu20253d,yu2025seqafford}, our approach leverages the rich knowledge embedded in VLMs to generalize to open-vocabulary understanding of affordances. 
This also enables the system to adapt to various intra-class variations (\eg, chairs with and without armrests) while maintaining consistent affordance identification across diverse object geometries and functional categories.

\begin{figure}[t]
    \centering
    \begin{subfigure}[b]{0.33\linewidth}
        \centering
        \includegraphics[width=\linewidth]{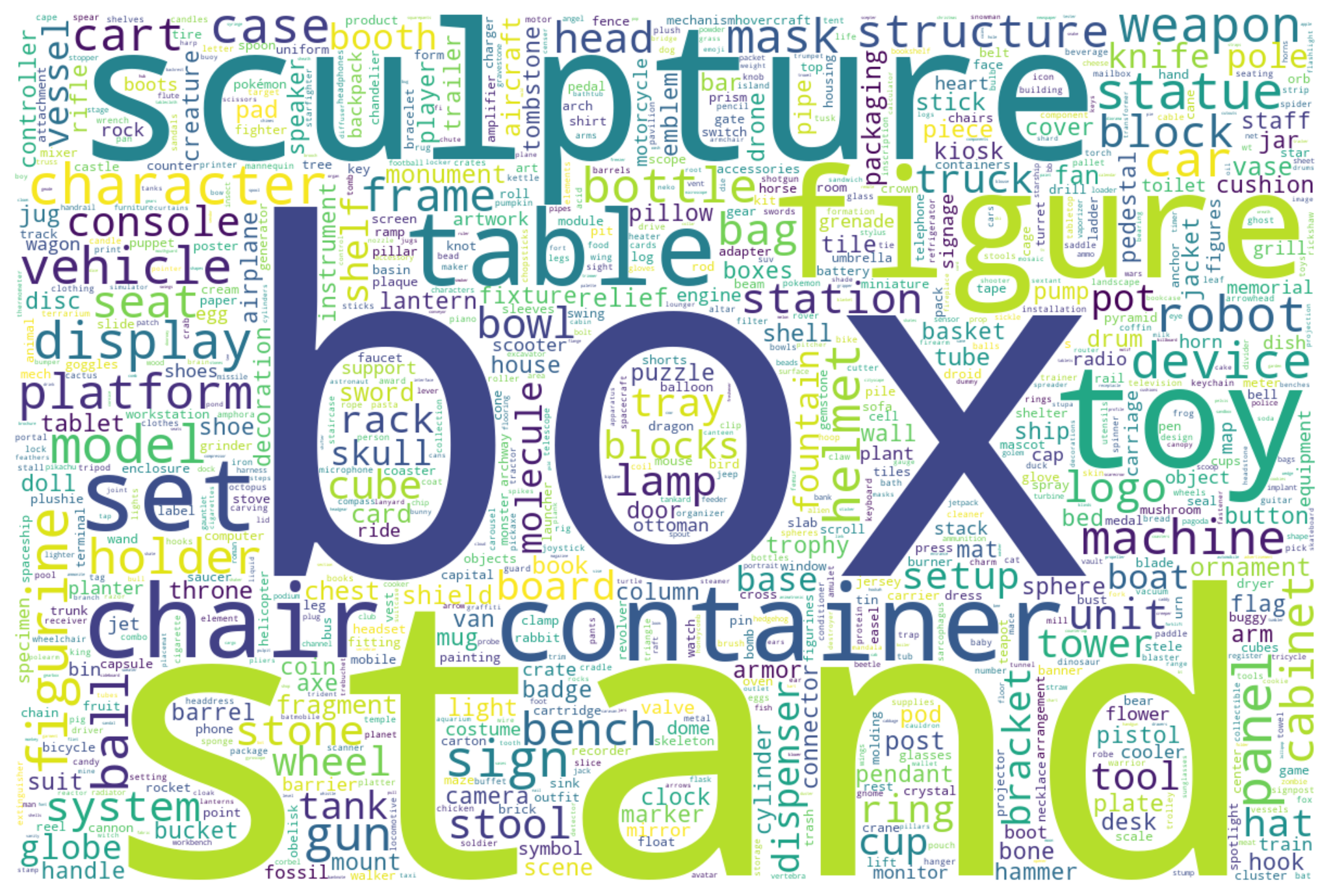}
        \caption{\small Object classes}  %
        \label{fig:wordcloud_object}
    \end{subfigure}
    \begin{subfigure}[b]{0.33\linewidth}
        \centering
        \includegraphics[width=\linewidth]{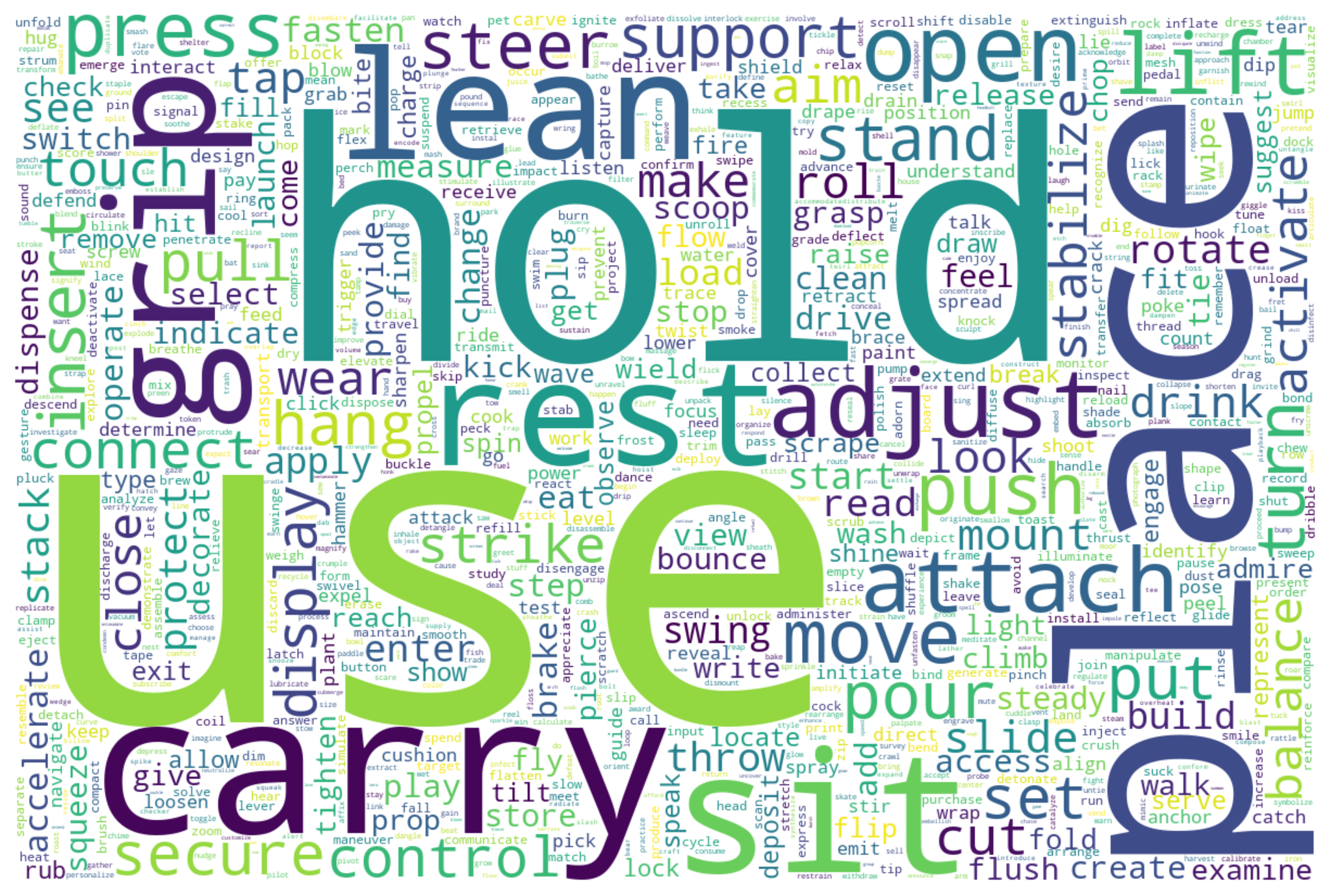}
        \caption{\small Affordance classes}  %
        \label{fig:wordcloud_affordance}
    \end{subfigure}
    \begin{subfigure}[b]{0.32\linewidth}
        \centering
        \includegraphics[width=\linewidth]{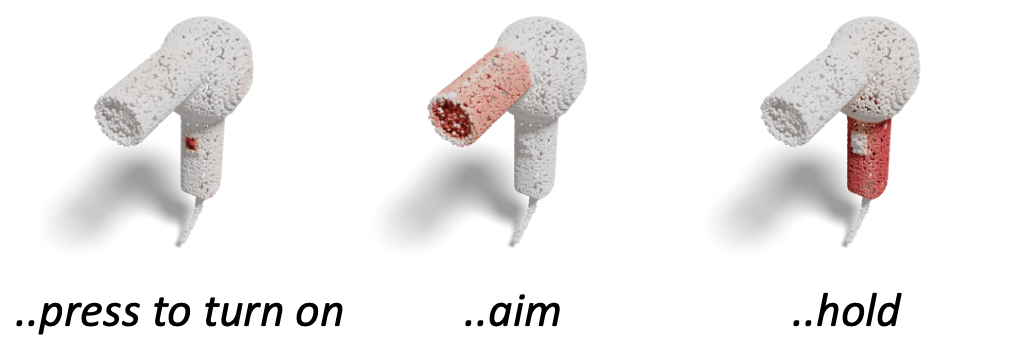}
        \caption{\small Diverse annotations}  %
        \label{fig:characteristics_diversity}
        \begin{minipage}[b]{\linewidth}
        \centering
        \scalebox{0.7}{
        \begin{tabular}{lrr}
            \toprule
            Dataset & Cov.~$\uparrow$ & Div.~$\uparrow$ \\
            \midrule
            LASO~\cite{li2024laso} & 0.6384 & 0.6578 \\
            \dataname & \textbf{0.7532} & \textbf{2.6638} \\
            \bottomrule
        \end{tabular}
        }
        \caption{\small Heatmap quality}
        \label{tab:characteristics_heatmap}
    \end{minipage}
    \end{subfigure}
    \caption{\textbf{Characteristics of the \dataname dataset.} (Left) The \dataname exhibit high diversity in both object and affordance categories. (Right) Quantitative comparison of heatmap coverage and diversity with LASO, highlighting that \dataname achieves significantly higher spatial coverage and distributional diversity (see Section~\ref{subsec:statistics} for details).}
    \label{fig:affogato_characteristics}
\end{figure}

\noindentbold{Stage 2: Language-guided interaction point prediction}
Once the affordance queries are generated, we utilize Molmo~\cite{deitke2025molmo}, a multimodal model capable of grounding natural language queries to spatial locations in images.
Specifically, Molmo demonstrates remarkable precision in identifying exact pixel locations when provided with input images and natural language prompts that request localization of specific regions in the image. 
This capability is crucial for accurately mapping affordance queries to their corresponding spatial locations on the object.
For each query and image pair, we instruct Molmo to predict pixel coordinates that represent the most likely interaction point in the given view.
The predicted interaction points across views are then used to guide the next stage.

\noindentbold{Stage 3: Affordance heatmap generation and aggregation}
We convert discrete interaction points into continuous 3D heatmap representations. First, the predicted point from each multi-view image is used as a prompt for MobileSAM~\cite{mobile_sam} to generate a 2D segmentation mask. To prioritize the precise interaction region over broader contextual areas, we adopt a smallest-mask selection strategy.
The segmentation logits are transformed through a sigmoid function to create probabilistic heatmaps with values between 0 and 1, representing the likelihood of an affordance at each pixel location.
These 2D heatmaps from multiple viewpoints are then projected onto the 3D object surface using the known camera parameters and depth information. 
We employ a voting-based aggregation process where each view contributes to the final 3D representation, with regions consistently identified across multiple views receiving higher confidence scores.
The output of this stage serves as our final 3D affordance heatmap annotations.

\noindentbold{Rendering 2D affordance heatmaps}
To bridge our 3D dataset with 2D image domains, we project the aggregated 3D affordance heatmaps onto 2D image planes from multiple viewpoints. 
Since the multi-view aggregation in Stage 3 resolves the inconsistency of Molmo and MobileSAM predictions, these rendered 2D heatmaps exhibit high consistency across viewpoints with accurate affordance regions.
For each object, we render heatmaps from 25 evenly distributed viewpoints and calculate affordance region visibility by summing the projected heatmap values. 
We select the most visible viewpoint for optimal representation.

\noindentbold{Error mitigation}
During the automatic dataset generation pipeline, we employ several strategies to mitigate errors. First, to enhance the fidelity of Gemma3’s responses, we adopt chain-of-thought (CoT) prompting~\cite{wei2022chain}, instructing the model to first predict the object’s semantic class and then generate affordance queries conditioned on that class and its functionalities. This approach prevents the model from relying solely on object shape, which often leads to affordance queries that fail to capture the object’s intended functions. Second, we apply a multi-view aggregation described in Stage 3 to reduce errors. Despite Molmo's exceptional pointing capabilities, its predicted interaction points can still contain errors and vary across viewpoints. Although a few views produce incorrect heatmaps, the consensus aggregation reinforces the majority of correct predictions, resulting in robust, view-consistent outputs.

\subsection{Statistics and analysis}
\label{subsec:statistics}

\begin{figure}[t]
    \centering
    \begin{minipage}[t]{0.35\linewidth}
        \centering
        \includegraphics[width=\linewidth]{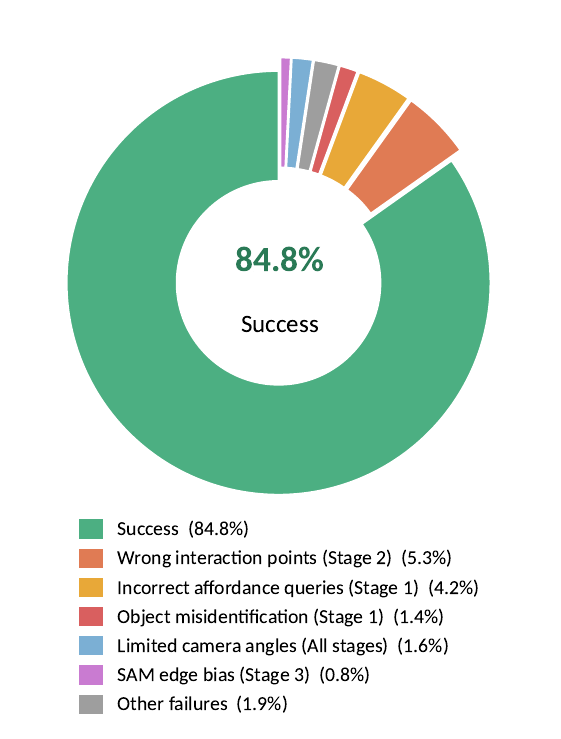}
    \end{minipage}
    \hfill
    \begin{minipage}[t]{0.63\linewidth}
        \centering
        \includegraphics[width=\linewidth]{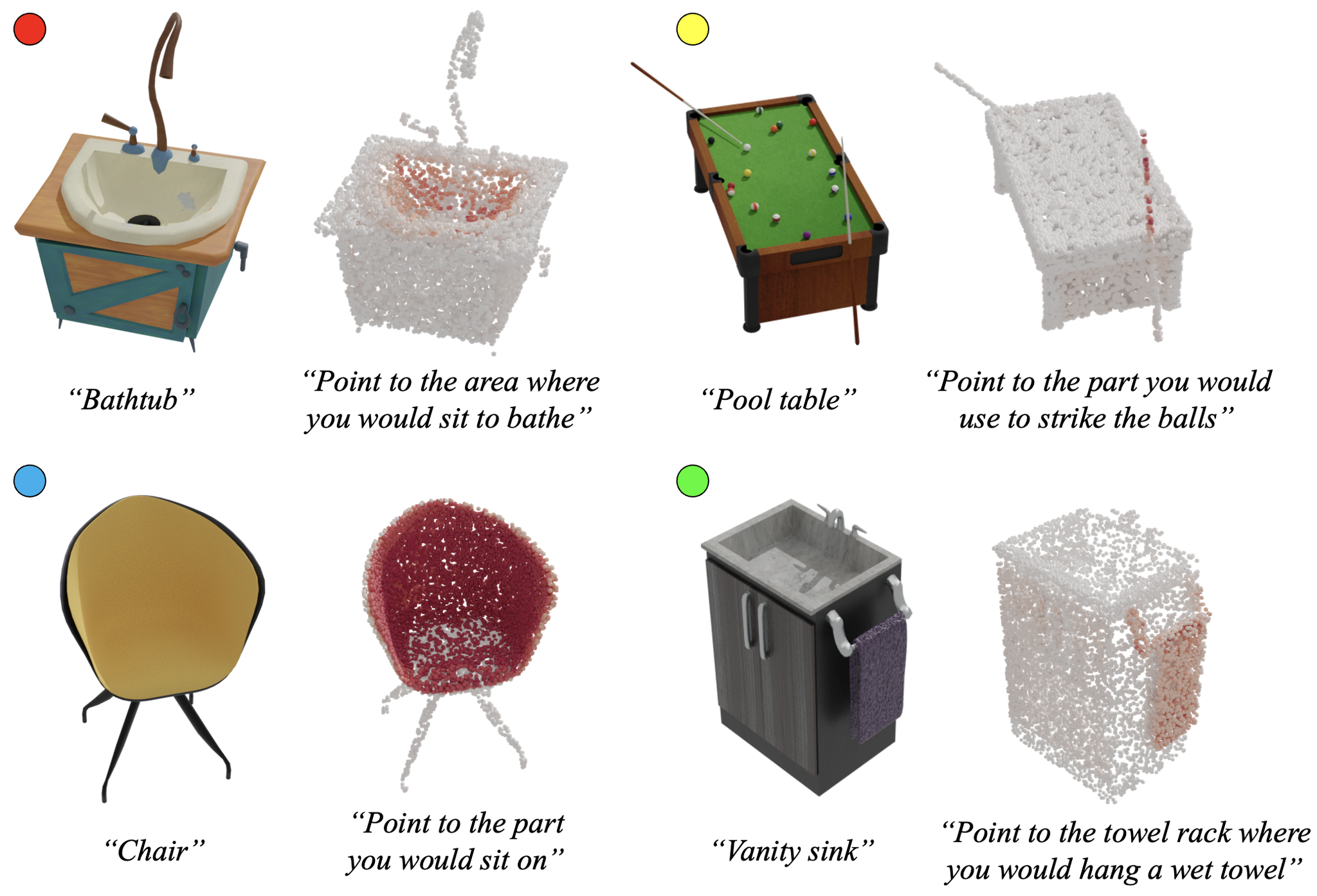}
    \end{minipage}
    \caption{
        (Left) Human validation results from \Secref{subsec:statistics}.
        (Right) Common failure cases in \dataname:
        \ccircle{red}~incorrect affordance queries from object misidentification;
        \ccircle{yellow}~partial coverage of the target affordance region;
        \ccircle{cyan}~SAM edge bias merging functionally distinct parts (\eg, backrest and seat);
        \ccircle{green}~failure in fine-grained grounding of the target object.
    }
    \label{fig:failure_piechart}
\end{figure}

\noindentbold{Data statistics}
As summarized in \Tref{tab:dataset_comparison}, \dataname provides over 750K affordance annotations across 150K 3D object instances in four categories: \textit{Daily-Used} includes the largest number of instances (121,799), followed by \textit{Transportations} (11,609), \textit{Furnitures} (8,759), and \textit{Electronics} (7,937), each with 5 affordance query-heatmap pairs.

\noindentbold{Semantic and spatial diversity}
Unlike existing datasets constrained by predefined taxonomies, \dataname spans a wide range of objects and affordances by leveraging VLM knowledge to generate diverse and context-aware affordance queries.
Figures~\ref{fig:wordcloud_object}~and~\ref{fig:wordcloud_affordance} demonstrate this semantic breadth across object classes and affordance types.
Our annotations capture varied interaction patterns, from precise point interactions (\eg button pressing) to extended surface interactions (\eg holding), as shown in Figure~\ref{fig:characteristics_diversity}.
Quantitatively, \Tref{tab:characteristics_heatmap} shows \dataname achieves significantly higher coverage (ratio of points covered by annotation union) and diversity (average pairwise KL divergence between heatmaps) scores than LASO~\cite{li2024laso}, indicating richer and more complementary affordance representations.

\noindentbold{Human validation and test split construction} 
To construct a reliable test split, we randomly sample 5K affordance query-heatmap pairs and assign them in disjoint subsets to 10 hired annotators for verification. We instruct annotators to rate each pair on semantic relevance, spatial accuracy, and heatmap coverage, and to inspect all stage-wise intermediates to locate the originating stage of any error.
This evaluation shows strong agreement with human judgment, achieving an 84.8\% success rate, and reveals the common failure modes of our pipeline (\Figref{fig:failure_piechart}).
We then build the test split from these verified pairs: incorrect queries are removed and incorrect interaction points are re-annotated through a dedicated annotation interface, yielding an object-instance-level test pool that is held out from training with no instance, view, or query overlap.

\section{\modelname models}
Our goal is to isolate the contribution of \dataname, attributing improvements in affordance grounding to the dataset itself rather than to model design.
However, prior 2D~\cite{agd20k,li2024one} and 3D~\cite{deng20213d,li2024laso,chu20253d} affordance methods have evolved in separate research streams, each accumulating its own modality-specific design choices.
Adopting such methods as cross-modal baselines would conflate architectural differences with dataset contribution, making it difficult to cleanly attribute performance gains to \dataname.
We therefore present \modelname, a unified baseline that is symmetric across 2D and 3D, isolating the dataset's contribution from architectural differences.

\begin{figure}[t!]
    \centering
    \includegraphics[width=\linewidth]{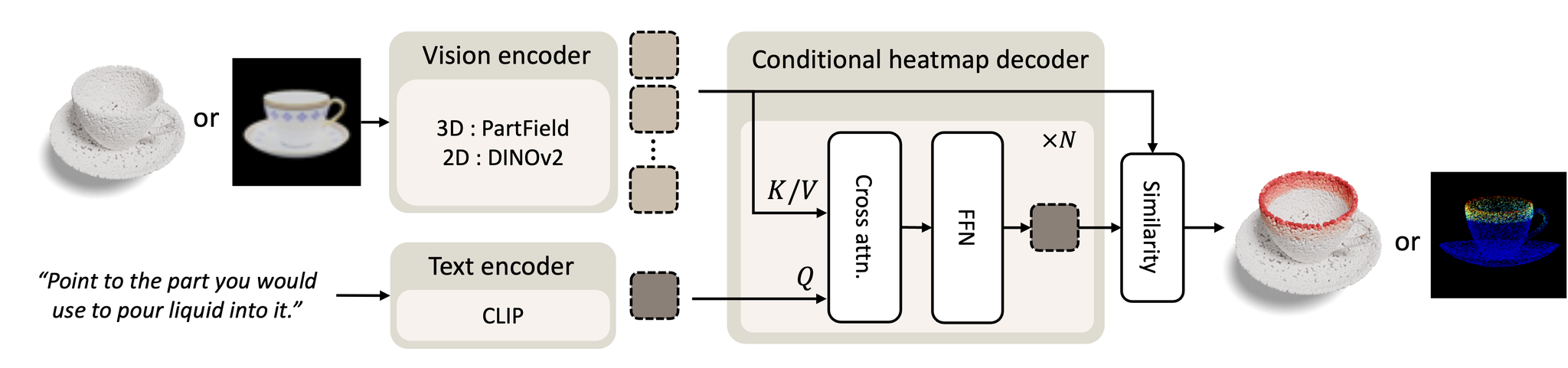}
    \caption{\textbf{\modelname-3D and \modelname-2D architectures.}
    Each model consists of a 3D or 2D visual encoder, a text encoder, and a text-conditioned heatmap decoder. 
    The affordance heatmap is predicted as the cross-modal similarity of the vision representation and the conditioned text embedding.
    } 
    \label{fig:architecture}
\end{figure}

Concretely, we present \modelname-3D for 3D point clouds and \modelname-2D for 2D images.
Each model consists of a modality-specific visual encoder, a text encoder, and a text-conditioned heatmap decoder (Figure~\ref{fig:architecture}).
The core of our design is a text-conditioned heatmap decoder that uses text embeddings as its queries instead of learnable ones.
Although this follows the standard transformer-based mask decoder architecture~\cite{cheng2021per,cheng2022masked,jain2023oneformer,schult2023mask3d,kolodiazhnyi2024oneformer3d}, using text embeddings as queries naturally supports open-vocabulary affordance grounding without predefined categories.
The following describes the modality-specific designs.

\noindentbold{\modelname-3D}
For the 3D vision encoder, we use the pretrained PartField model~\cite{liu2025partfield}, which captures generic part concepts from large-scale 3D data.
For the text encoder, we employ Recap-CLIP~\cite{li2024if}, which provides robust language understanding capabilities.
Following PointRefer~\cite{li2024laso}, we finetune both the pretrained 3D vision encoder and text encoder during training to adapt them to our downstream task while leveraging their strong pretrained representations.

\noindentbold{\modelname-2D}
For the 2D vision encoder, we use the pretrained DINOv2 model~\cite{dinov2}, and CLIP~\cite{clip} as the text encoder.
Following OOAL~\cite{li2024one}, we freeze both encoders during training to maintain their pretrained representations.
Our architecture adapts the design of OOAL~\cite{li2024one} but differs in that we remove the learnable prompt component and take a single affordance query as input, following the referring expression segmentation framework.

\section{Experiments}
\subsection{Datasets and baselines}
\noindentbold{3D datasets}
We primarily evaluate \modelname-3D on the LASO dataset~\cite{li2024laso}, a benchmark for open-vocabulary 3D affordance grounding with free-form text queries.
LASO provides both seen and unseen settings; the seen setting evaluates on object/affordance classes observed during training, while the unseen setting tests on held-out classes to assess zero-shot generalization.
We further evaluate on our \dataname test split, a held-out set of 5K human-verified pairs that covers substantially more diverse object categories and affordance concepts than LASO.
It serves as a more challenging testbed and provides both in-domain and cross-domain settings to probe open-vocabulary generalization beyond the limited category coverage of existing benchmarks.

\noindentbold{2D datasets}
We evaluate \modelname-2D on the AGD20K~\cite{agd20k}, which consists of two object splits:
seen and unseen.
AGD20K-Weak, 1-shot, and Full represent weakly supervised, one-shot, and fully supervised versions of the AGD20K training dataset, respectively.

\noindentbold{3D baselines}
We select representative baselines for open-vocabulary 3D affordance grounding, which take language descriptions paired with 3D point clouds as input.
Our evaluation includes multiple representative methods: Referring Transformer~\cite{li2021referring}, 3D-SPS~\cite{luo20223d}, ReLA~\cite{liu2023gres}, IAGNet~\cite{yang2023grounding}, OpenAD~\cite{nguyen2023open}, and PointRefer~\cite{li2024laso}.
For OpenAD, since official results on LASO~\cite{li2024laso} were not available, we trained the model on the LASO dataset using their official code implementation.
All other baseline results are taken directly from~\cite{li2024laso}.
We compare these methods with our proposed \modelname-3D on both LASO seen and unseen splits to evaluate generalization performance.

\noindentbold{2D baselines}
For zero‑shot evaluation on AGD20K, 
we use Molmo+SAM2~\cite{deitke2025molmo, ravi2025sam}, LISA-7B~\cite{lisa} (a reasoning segmentation model), M$^2$SA-7B~\cite{jang2025mmr} (a part-level referring segmentation model), and AffordanceNet~\cite{wu2025ragnet} (a reasoning-based affordance segmentation model).
We format the open-vocabulary query for an affordance as ``\textit{Point to the part that you should interact with to \{affordance\}}''.
For the affordance-specific model baselines in Table~\ref{tab:agd20k-all} (b), we adopt results from \cite{affordancellm} and additionally include AffordanceNet~\cite{wu2025ragnet}.

\noindentbold{Evaluation metrics}
For 3D evaluation, following prior work~\cite{li2024laso}, we use average Intersection over Union (aIoU), Area Under the ROC Curve (AUC), Similarity (SIM), and Mean Absolute Error (MAE).
Since MAE is sensitive to annotation scale, we primarily report aIoU, AUC, and SIM.
For 2D evaluation, we use Kullback-Leibler Divergence (KLD), Similarity (SIM), and Normalized Scanpath Saliency (NSS)~\cite{bylinskii2019different}.
Higher values are better for aIoU, AUC, SIM, and NSS while lower values are better for MAE and KLD.

\def\Dg#1{\raisebox{-.5ex}{\tiny\textcolor{green!60!black}{$#1$}}}%
\def\Dr#1{\raisebox{-.5ex}{\tiny\textcolor{gray!70!black}{$#1$}}}%
\begin{table}[t]
    \centering
    \caption{
        Open-vocabulary 3D affordance grounding on the LASO~\cite{li2024laso} test split. ``\dataname pretrain'' denotes pretraining on the \dataname train split and then finetuned on the LASO train split. Subscript values indicate the effect of \dataname pretraining (green: improvement, gray: degradation).
    } 
    \label{tab:laso}
    \resizebox{\linewidth}{!}{
        \begin{tabular}{l|llll|llll}
            \toprule
            \multirow{2}{*}{Method} & \multicolumn{4}{c|}{Seen} & \multicolumn{4}{c}{Unseen} \\
            & aIoU\hspace{0.5pt}$\uparrow$ & AUC\hspace{0.5pt}$\uparrow$ & SIM\hspace{0.5pt}$\uparrow$ & MAE\hspace{0.5pt}$\downarrow$ & aIoU\hspace{0.5pt}$\uparrow$ & AUC\hspace{0.5pt}$\uparrow$ & SIM\hspace{0.5pt}$\uparrow$ & MAE\hspace{0.5pt}$\downarrow$\\
            \midrule
            Ref. Trans.~\cite{li2021referring} & 13.7 & 79.8 & 49.7 & 0.124 & 10.2 & 69.1 & 43.2 & 0.145 \\
            3D-SPS~\cite{luo20223d} & 11.4 & 76.2 & 43.3 & 0.138 & ~~7.9 & 68.8 & 40.2 & 0.158 \\
            ReLA~\cite{liu2023gres} & 15.2 & 78.9 & 53.2 & 0.118 & 10.7 & 69.7 & 42.9 & 0.144 \\
            IAGNet~\cite{yang2023grounding} & 17.8 & 82.3 & 56.1 & 0.109 & 12.9 & 77.8 & 44.3 & 0.129 \\
            \midrule
            OpenAD~\cite{nguyen2023open} & 14.2 & 85.1 & 53.3 & 0.103 & 14.6 & 80.7 & 51.8 & 0.109 \\
            {\footnotesize $+$ \dataname pretrain} & 16.1\Dg{+1.9} & 86.8\Dg{+1.7} & 53.9\Dg{+0.6} & 0.100\Dg{-.003} & 15.5\Dg{+0.9} & 81.8\Dg{+1.1} & 53.4\Dg{+1.6} & 0.103\Dg{-.006} \\
            \midrule
            PointRefer~\cite{li2024laso} & 20.8 & \textbf{87.3} & 62.9 & \textbf{0.093} & 14.6 & 80.2 & 50.7 & 0.119 \\
            {\footnotesize $+$ \dataname pretrain}   & 20.2\Dr{-0.6} & 86.0\Dr{-1.3} & 60.0\Dr{-2.9} & 0.098\Dr{+.005} & 18.6\Dg{+4.0} & 81.4\Dg{+1.2} & 56.1\Dg{+5.4} & 0.103\Dg{-.016} \\
            \midrule
            \rowcolor{gray!15} \modelname-3D & 20.4 & 86.0 & 63.3 & 0.102 & 18.7 & 80.0 & 60.0 & \textbf{0.101} \\
            \rowcolor{gray!15} {\footnotesize $+$ \dataname pretrain}   & \textbf{21.9}\Dg{+1.5} & 85.9\Dr{-0.1} & \textbf{63.7}\Dg{+0.4} & 0.116\Dr{+.014} & \textbf{20.8}\Dg{+2.1} & \textbf{82.9}\Dg{+2.9} & \textbf{61.4}\Dg{+1.4} & 0.122\Dr{+.021} \\
            \bottomrule
        \end{tabular}
    }
\end{table}

\begin{figure}[t]
    \centering
    \includegraphics[width=0.85\linewidth]{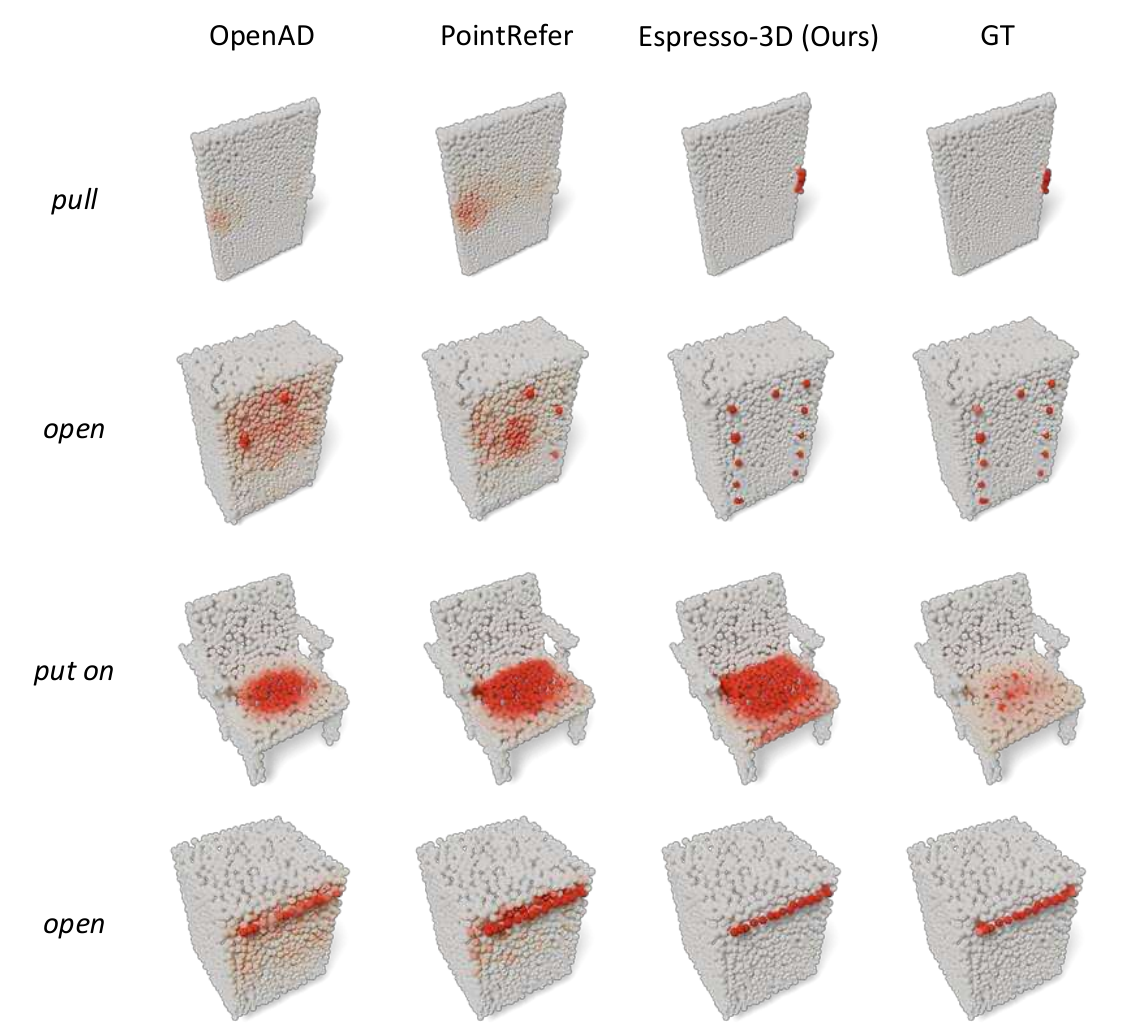}
    \caption{Qualitative comparison between OpenAD~\cite{nguyen2023open}, PointRefer~\cite{li2024laso}, and \modelname-3D on the LASO~\cite{li2024laso} test split.}
    \label{fig:qual3d}
\end{figure}

\begin{table}[t]
    \centering
    \caption{
    Open-vocabulary 3D affordance grounding on the \dataname test split. \textit{All $\rightarrow$ All} denotes in-domain evaluation, while \textit{Daily-used $\rightarrow$ Furnitures} and \textit{Furnitures $\rightarrow$ Daily-used} represent cross-domain settings.
}
    \label{tab:affogato_test}
    \resizebox{\linewidth}{!}{
        \begin{tabular}{l|cccc|cccc|cccc}
            \toprule
            \multirow{2}{*}{Method} & \multicolumn{4}{c|}{\textit{All $\rightarrow$ All}} & \multicolumn{4}{c|}{\textit{Daily-used $\rightarrow$ Furnitures}} & \multicolumn{4}{c}{\textit{Furnitures $\rightarrow$ Daily-used}} \\
            & aIoU\hspace{0.5pt}$\uparrow$ & AUC\hspace{0.5pt}$\uparrow$ & SIM\hspace{0.5pt}$\uparrow$ & MAE\hspace{0.5pt}$\downarrow$ & aIoU\hspace{0.5pt}$\uparrow$ & AUC\hspace{0.5pt}$\uparrow$ & SIM\hspace{0.5pt}$\uparrow$ & MAE\hspace{0.5pt}$\downarrow$ & aIoU\hspace{0.5pt}$\uparrow$ & AUC\hspace{0.5pt}$\uparrow$ & SIM\hspace{0.5pt}$\uparrow$ & MAE\hspace{0.5pt}$\downarrow$ \\
            \midrule
            OpenAD & ~~3.1 & 64.8 & 32.9 & 0.150 & ~~7.6 & 67.8 & 36.8 & 0.177 & ~~1.6 & 54.3 & \textbf{30.8} & 0.201 \\
            PointRefer & 10.5 & 76.1 & 40.5 & 0.120 & 13.5 & 78.9 & 44.3 & 0.157 & ~~2.8 & 60.6 & 26.5 & \textbf{0.122} \\
            \rowcolor{gray!15} \modelname-3D & \textbf{13.6} & \textbf{79.0} & \textbf{42.9} & \textbf{0.111} & \textbf{18.2} & \textbf{80.4} & \textbf{47.5} & \textbf{0.125} & \textbf{~~4.6} & \textbf{62.4} & 30.4 & 0.130 \\
            \bottomrule
        \end{tabular}
    }
\end{table}

\subsection{3D affordance grounding}
\noindentbold{LASO~\cite{li2024laso} test split}
Table~\ref{tab:laso} presents the 3D affordance grounding results on LASO.
Without pretraining on \dataname, all methods achieve reasonable performance on the seen setting but show limited generalization to unseen object-affordance combinations.
Pretraining on \dataname consistently and substantially improves unseen performance across all methods: OpenAD gains 0.9\%p aIoU, PointRefer achieves the largest gain of 4.0\%p aIoU (18.6 vs. 14.6), and \modelname-3D gains 2.1\%p aIoU.
While pretrained PointRefer shows a slight degradation on the seen setting (20.2 vs. 20.8 aIoU), it achieves the largest unseen improvement; \modelname-3D improves consistently on both seen (+1.5\%p aIoU) and unseen (+2.1\%p aIoU).
These results confirm that \dataname pretraining provides a strong and consistent benefit for open-vocabulary generalization, regardless of model architecture.
Figure~\ref{fig:qual3d} provides a qualitative comparison on the LASO test split.

\noindentbold{\dataname test split}
We evaluate open-vocabulary 3D affordance grounding on the \dataname test split under both in-domain and cross-domain settings (Table~\ref{tab:affogato_test}), comparing \modelname-3D against OpenAD~\cite{nguyen2023open} and PointRefer~\cite{li2024laso} retrained on our dataset.
Since benchmarks like LASO cover few categories, this experiment tests generalization under the broader and more diverse distribution of \dataname.
The in-domain split (\textit{All $\rightarrow$ All}) trains and evaluates on all objects, while the cross-domain splits (\textit{Daily-used $\rightarrow$ Furnitures} and \textit{Furnitures $\rightarrow$ Daily-used}) train on one object domain and test on the other to measure cross-domain transfer.
Even the strongest methods leave substantial headroom on \dataname, reflecting the broad object and affordance coverage over which it requires generalization.
Across all three settings, \modelname-3D consistently outperforms OpenAD and PointRefer, mirroring their ranking on LASO and indicating that these trends are not benchmark-specific.
The cross-domain results are strongly asymmetric: training on diverse daily-used objects and testing on furniture (\textit{Daily-used $\rightarrow$ Furnitures}) far outperforms the reverse (18.2 vs.\ 4.6 aIoU for \modelname-3D), since the narrow furniture domain covers little of the broader daily-used categories.

\subsection{2D affordance grounding}
\begin{figure}[t!]
    \centering
    \includegraphics[width=\linewidth]{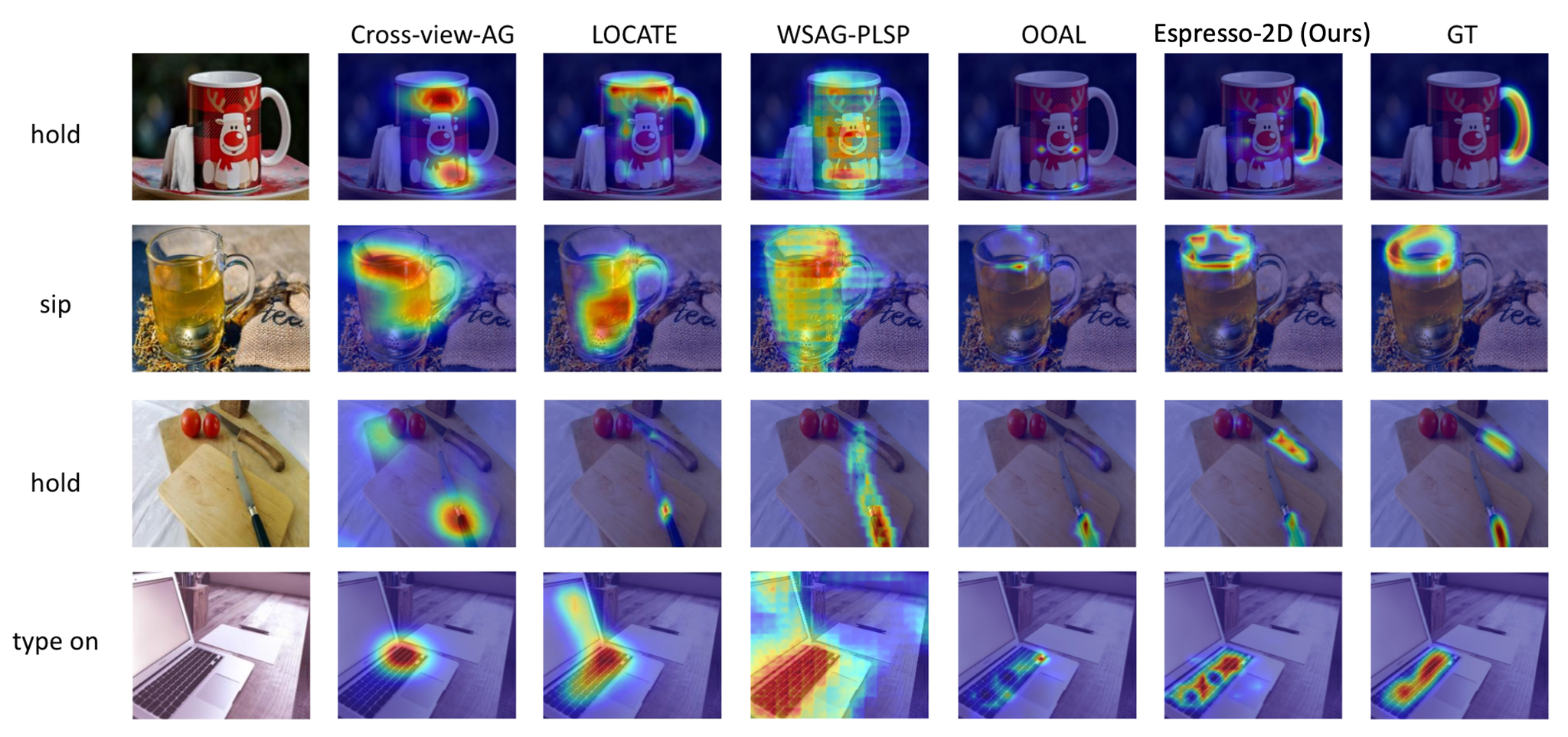}
    \vspace{-2mm}
    \caption{Qualitative comparison between Cross-view-AG~\cite{agd20k}, LOCATE~\cite{li2023locate}, WSAG-PLSP~\cite{xuweakly}, OOAL~\cite{li2024one}, and \modelname-2D, which is pretrained on the \dataname train split and then finetuned on the AGD20K train split with full supervision.}
    \label{fig:qual2d}
\end{figure}

Table~\ref{tab:agd20k-all} (a) presents results on the test sets of AGD20K's seen and unseen object splits, evaluated zero-shot without any AGD20K training. Each baseline is pretrained on its own respective dataset, while \modelname-2D is pretrained on \dataname.
\modelname-2D with \dataname pretraining outperforms all baselines across both test sets, achieving 40.2 SIM on the seen split and 37.6 SIM on the unseen split versus 22 to 30 SIM for competing methods, demonstrating that \dataname pretraining yields affordance representations that generalize well across diverse object categories.
Table~\ref{tab:agd20k-all} (b) presents supervised learning results across different supervision levels.
\modelname-2D with \dataname pretraining, fine-tuned on AGD20K, achieves the best performance across all supervision settings.
Notably, \dataname pretraining provides consistent improvements over training solely on AGD20K---the \dataname-pretrained \modelname-2D outperforms its from-scratch counterpart by 0.060 KLD and 1.6 SIM under full supervision, and AffordanceNet similarly benefits from \dataname pretraining---validating that \dataname’s diverse affordance concepts transfer across domain boundaries even when target-domain data is available.
Figure~\ref{fig:qual2d} shows qualitative examples.

\begin{table}[!t]
    \caption{Open-vocabulary 2D affordance grounding on the AGD20K~\cite{agd20k} test split. ``\dataname pretrain'' denotes pretraining on the \dataname train split.
    }
    \vspace{-4mm}
    \label{tab:agd20k-all}
    \begin{subtable}[t]{.47\linewidth}
    \centering
    \caption{Zero-shot results} \label{tab:agd20k-zeroshot}
    \vspace{-3mm}
    \resizebox{\linewidth}{!}{
        \begin{tabular}{l|ccc}
            \toprule
            Method & KLD\hspace{0.5pt}$\downarrow$ & SIM\hspace{0.5pt}$\uparrow$ & NSS\hspace{0.5pt}$\uparrow$ \\
            \midrule
            \textit{Seen split} & \\
            Molmo+SAM2\cite{deitke2025molmo, ravi2025sam} & 1.804  &26.1 &0.729\\
            LISA-7B~\cite{lisa} & 1.627 & 29.6 & 0.819  \\
            M$^2$SA-7B~\cite{jang2025mmr} & 1.772 & 25.8 & 0.620 \\
            AffordanceNet~\cite{wu2025ragnet} & 1.729 & 26.6 & 0.755 \\
            \rowcolor{gray!15} \modelname-2D & \textbf{1.426} & \textbf{40.2} & \textbf{0.985} \\ \midrule
            \textit{Unseen split} & \\
            Molmo+SAM2~\cite{deitke2025molmo, ravi2025sam} & 1.953  & 22.6 & 0.718\\
            LISA-7B~\cite{lisa} & 1.830 & 25.6 & 0.765 \\
            M$^2$SA-7B~\cite{jang2025mmr} & 1.925 & 22.7 & 0.657 \\
            AffordanceNet~\cite{wu2025ragnet} & 1.932 & 22.6 & 0.700 \\
            \rowcolor{gray!15} \modelname-2D & \textbf{1.571} & \textbf{37.6} & \textbf{1.016} \\
            \bottomrule
        \end{tabular}
    }
    \end{subtable}%
    \hfill
    \begin{subtable}[t]{.53\linewidth}
    \centering
    \caption{Supervised learning results}
    \label{tab:agd20k-train}
    \vspace{-3mm}
    \resizebox{\linewidth}{!}{
        \begin{tabular}{ll|ccc}
            \toprule
             Method & Sup. & KLD\hspace{0.5pt}$\downarrow$ & SIM\hspace{0.5pt}$\uparrow$ & NSS\hspace{0.5pt}$\uparrow$ \\
            \midrule
            Cross-view-AG~\cite{agd20k}           & \multirow{5}{*}{Weak} & 1.787 & 28.5 & 0.829 \\
            Cross-view-AG+~\cite{luo2024grounded} &                      & 1.765 & 27.9 & 0.882 \\
            AffCorrs~\cite{hadjivelichkov2023one} &                      & 1.618 & 34.8 & 1.021 \\
            LOCATE~\cite{li2023locate}            &                      & 1.405 & 37.2 & 1.157 \\
            WSAG-PLSP~\cite{xuweakly}             &                      & 1.153 & 43.7 & 1.418 \\
            \midrule
            OOAL~\cite{li2024one} & \multirow{1}{*}{1-shot} & 1.070 & 46.1 & 1.503 \\
            \midrule
            LOCATE-Sup~\cite{li2023locate}      &\multirow{6}{*}{Full} & 1.907 & 23.6 & 0.641 \\
            LOCATE-Sup-OWL~\cite{li2023locate, minderer2022simple} & & 1.927 & 23.4 & 0.624\\
            AffordanceLLM~\cite{affordancellm} & & 1.463 & 37.7 & 1.070 \\
            AffordanceNet~\cite{wu2025ragnet} & & 1.871 & 30.5 & 0.750 \\
            {\footnotesize $+$ \dataname pretrain} & & 1.721 & 33.7 & 0.852 \\
            \midrule
            \cellcolor{gray!15}\modelname-2D & \cellcolor{gray!15} & \cellcolor{gray!15}1.034  & \cellcolor{gray!15}50.3 & \cellcolor{gray!15}1.550 \\
            \rowcolor{gray!15} {\footnotesize $+$ \dataname pretrain} & & \textbf{0.974} & \textbf{51.9} & \textbf{1.645}  \\
            \bottomrule
        \end{tabular}
    }
    \end{subtable}
\end{table}

\subsection{Analysis}

\begin{table*}[t]
    \centering
    \caption{
        Training-evaluation overlap analysis on LASO unseen and AGD20K unseen.
    }
    \vspace{-4mm}
    \label{tab:label_leakage}
    \begin{subtable}[t]{.46\linewidth}
        \centering
        \caption{LASO unseen split (3D)}
        \vspace{-3mm}
        \scalebox{0.9}{
            \begin{tabular}{cc|cccc}
                \toprule
                Pretraining & Filtering & aIoU\hspace{0.5pt}$\uparrow$ & AUC\hspace{0.5pt}$\uparrow$ & SIM\hspace{0.5pt}$\uparrow$ & MAE\hspace{0.5pt}$\downarrow$ \\
                \midrule
                & & 18.7 & 80.0 & 60.0 & 0.101 \\
                \checkmark & & 20.8 & 82.9 & 61.4 & 0.122 \\
                \rowcolor{gray!15} \checkmark & \checkmark & 20.2 & 82.4 & 61.1 & 0.119 \\
                \bottomrule
            \end{tabular}
        }
    \end{subtable}
    \hfill
    \begin{subtable}[t]{.46\linewidth}
        \centering
        \caption{AGD20K unseen split (2D)}
        \vspace{-3mm}
        \scalebox{0.9}{
            \begin{tabular}{cc|ccc}
                \toprule
                Pretraining & Filtering & KLD\hspace{0.5pt}$\downarrow$ & SIM\hspace{0.5pt}$\uparrow$ & NSS\hspace{0.5pt}$\uparrow$ \\
                \midrule
                & & 1.034 & 50.3 & 1.550 \\
                \checkmark & & 0.974 & 51.9 & 1.645 \\
                \rowcolor{gray!15} \checkmark & \checkmark & 1.016 & 52.0 & 1.631 \\
                \bottomrule
            \end{tabular}
        }
    \end{subtable}
\end{table*}

\noindentbold{Effect of held-out category filtering}
To check that the pretraining gains are not simply due to \dataname covering categories that each benchmark holds out, we remove the pretraining samples that fall into those held-out classes and re-train.
We query Gemma3~\cite{team2025gemma} to judge whether each sample's class label falls into a benchmark's held-out categories, and conservatively remove any potential match. For LASO, which holds out (object, affordance) pairs, we drop samples whose predicted pair appears in the evaluation set (5.46\%); for AGD20K, which holds out object categories, we exclude samples assigned to those categories (14.0\%). Since these predictions are imperfect, the filter likely over-removes, making the numbers a lower bound.
Even so, filtered pretraining consistently surpasses training from scratch on both benchmarks (Table~\ref{tab:label_leakage}; on LASO, 20.2 vs.\ 18.7 aIoU, within 0.6 of full pretraining), confirming that \dataname's benefits come from diverse supervision at scale rather than from covering the held-out categories.

\begin{figure}[t]
    \centering
    \includegraphics[width=0.95\linewidth]{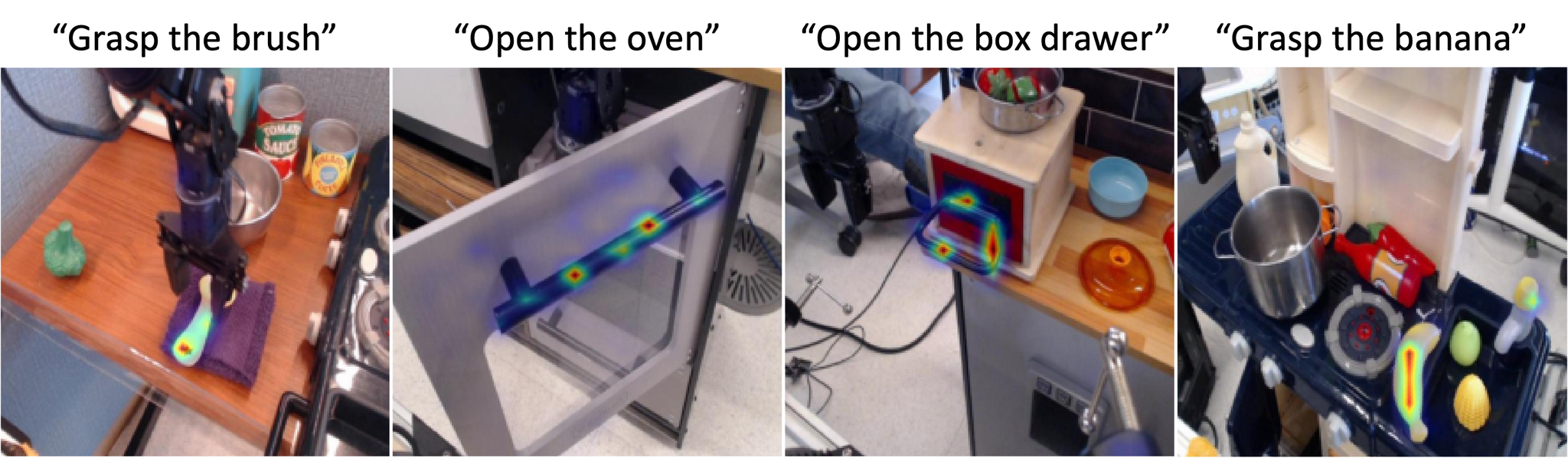}
    \vspace{-2mm}
    \caption{\textbf{Zero-shot \modelname-2D on real-world robot scenes.} Despite the visual domain gap from Objaverse renderings, \modelname-2D pretrained on \dataname localizes open-vocabulary affordances correctly across diverse, cluttered Open X-Embodiment~\cite{o2024open} robot workspace images.}
    \label{fig:oxe}
\end{figure}

\noindentbold{Synthetic-to-real generalization}
Although \dataname is built from clean, isolated Objaverse renderings, the pretrained \modelname-2D transfers to realistic, in-the-wild inputs. To probe a large domain gap, we evaluate \modelname-2D zero-shot on cluttered robot workspace images from Open X-Embodiment~\cite{o2024open}. As shown in Figure~\ref{fig:oxe}, despite never observing such scenes during training, \modelname-2D still localizes the queried affordances correctly across diverse robot-manipulation scenes, indicating that the affordance representations learned from \dataname generalize beyond the synthetic source domain.

\begin{figure*}[t]
    \centering
    \begin{subfigure}[t]{0.66\textwidth}
        \centering
        \includegraphics[width=\linewidth]{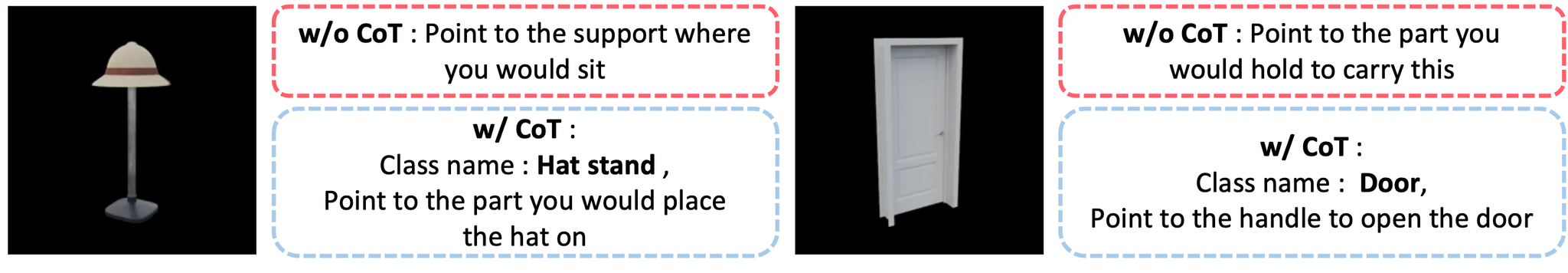}
        \caption{CoT prompting}
        \label{fig:cot}
    \end{subfigure}
    \hfill
    \begin{subfigure}[t]{0.30\textwidth}
        \centering
        \includegraphics[width=1.0\linewidth]{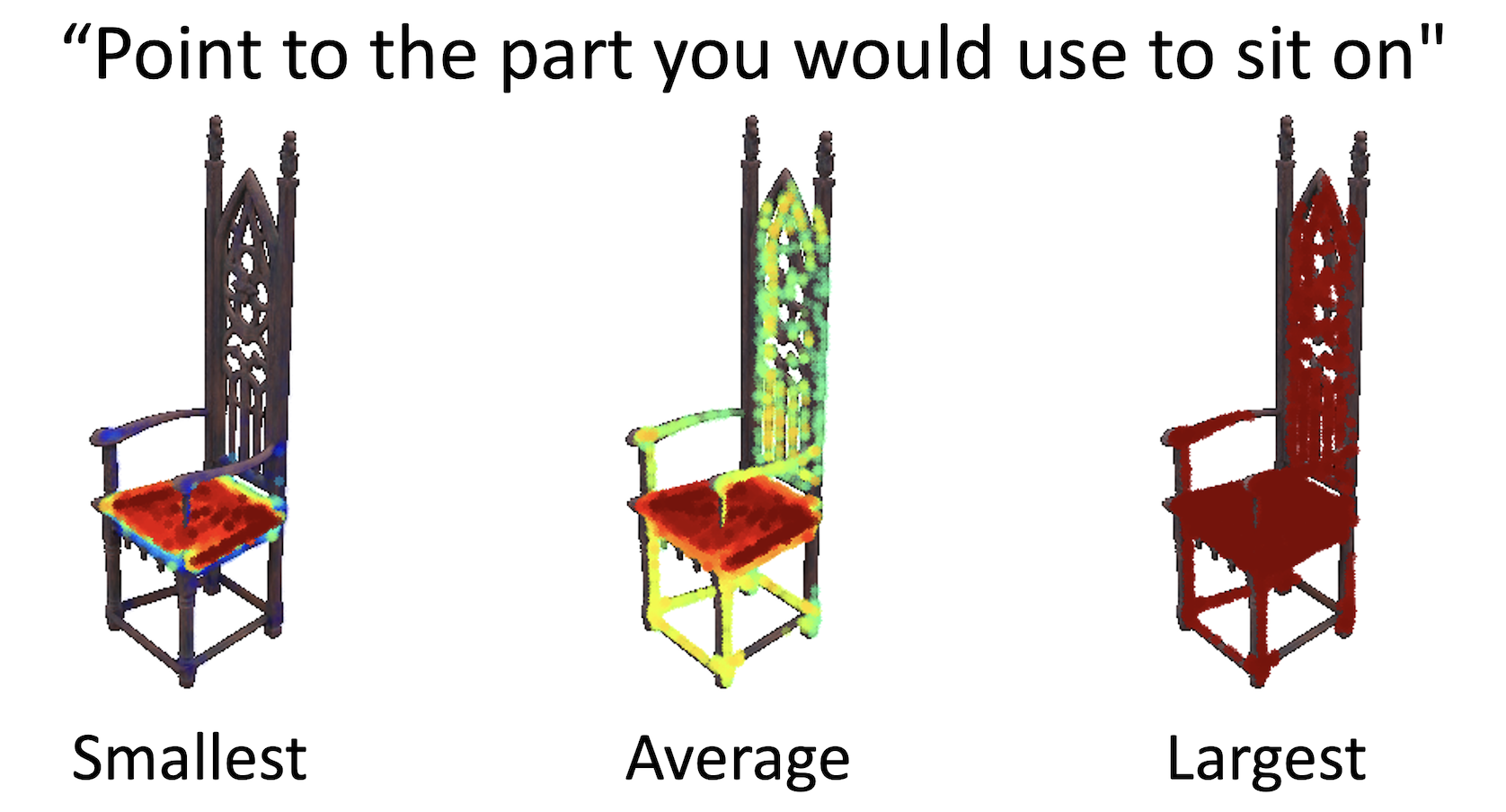}
        \caption{Mask-size selection}
        \label{fig:mask_size}
    \end{subfigure}
    \vspace{-3mm}
    \caption{\textbf{Analyses of our data annotation pipeline.} (a) CoT prompting produces functionally grounded affordance queries rather than shape-only guesses. (b) Among the three mask hypotheses returned by SAM, the smallest isolates part-level structure at affordance granularity.}
    \label{fig:pipeline_analyses}
\end{figure*}

\noindentbold{Effect of chain-of-thought prompting}
We qualitatively analyze the effect of CoT~\cite{wei2022chain} prompting in our data annotation pipeline for affordance query generation. As shown in Figure~\ref{fig:cot}, without CoT prompting the model often fails to capture the object’s functional properties and instead relies primarily on shape when generating queries. For example, it produces “Point to the support where you would sit.” for a hat stand or “Point to the part you would hold to carry this.” for a door. In contrast, CoT prompting encourages functional reasoning, producing more appropriate queries such as “Point to the handle to open the door,” or “Point to the part you would place the hat on.”

\noindentbold{Effect of mask-size selection}
Another key design choice in our pipeline concerns how interaction points are turned into region masks.
SAM~\cite{mobile_sam,ravi2025sam} returns three mask hypotheses for each interaction-point prompt, and we adopt the \emph{smallest} one, as stated in Section~\ref{subsec:data_pipeline}. Figure~\ref{fig:mask_size} compares the three choices: the larger masks tend to snap to whole-object silhouettes, whereas the smallest mask isolates part-level structures (\eg, handle, lid, button) at affordance granularity. Although the smallest mask can occasionally miss parts in individual views, our multi-view aggregation closes such per-view gaps, making the smallest-mask choice the most reliable basis for precise affordance heatmaps.

\section{Discussion and conclusion}

We have presented \frameworkname, a unified framework for open-vocabulary affordance grounding across 3D and 2D domains. By chaining foundation models, our automated pipeline produces \dataname, the largest affordance grounding dataset with 750K affordance annotations across 150K 3D assets, eliminating costly manual annotation while capturing diverse interactions beyond predefined categories. Our \modelname-3D and \modelname-2D models consistently benefit from pretraining on \dataname, demonstrating improved generalization to unseen object-affordance combinations. Notably, our annotation pipeline can be readily applied to other 3D datasets, offering a practical path toward ever-growing affordance supervision. While our framework effectively localizes contact points for a given action, predicting finer-grained information such as temporal dynamics and force requirements~\cite{kim2024beyond} remains an open challenge. We believe \frameworkname and \dataname provide a solid foundation for future research toward more comprehensive embodied AI systems.

\section*{Acknowledgements}

This work was supported by RLWRLD Inc. and also by IITP grants (RS-2022-II220959: Few-Shot Learning of Causal Inference in Vision and Language for Decision Making (50\%), RS-2022-II220290: Visual Intelligence for Space-Time Understanding \& Generation (25\%), RS-2024-00457882: National AI Research Lab Project (20\%), RS-2019-II191906: AI Graduate School Program at POSTECH (5\%)) funded by Ministry of Science and ICT, Korea.

\bibliographystyle{splncs04}
\bibliography{main}

\clearpage

\appendix
\renewcommand{\thesection}{\Alph{section}}
\renewcommand{\thetable}{A\arabic{table}}
\renewcommand{\thefigure}{A\arabic{figure}}
\setcounter{section}{0}
\setcounter{table}{0}
\setcounter{figure}{0}
\section{Dataset link}
\href{https://huggingface.co/datasets/project-affogato/affogato}{https://huggingface.co/datasets/project-affogato/affogato}

\section{Implementation details}
\label{sec:implementation_detail}

\subsection{Data Annotation Pipeline}
\label{subsec:data_annotation_detail}
\noindentbold{Use of pretrained models}
Our data annotation pipeline uses the following pretrained models: (1) Gemma3~\cite{team2025gemma} (\url{google/gemma-3-4b-it}) for generating natural language affordance queries
(2) Molmo~\cite{deitke2025molmo} (\url{allenai/Molmo-7B-D-0924}) for predicting interaction points in 2D images, 
and (3) MobileSAM~\cite{mobile_sam} for predicting 2D heatmap given interaction point prompts.

\noindentbold{Computational resources}
For \dataname dataset generation, we use 8 NVIDIA H100 GPUs with 80GB of memory each. The data generation pipeline takes approximately 24 hours to process 150K Objaverse instances.

\noindentbold{Image sampling}
G-Objaverse~\cite{zuo2024sparse3d} provides 38 views per object. For computational efficiency, we used the first 25 views that are captured with the same elevation but uniformly distributed azimuths around the object.
For stage 1 of our pipeline, we sample 5 images at equal intervals from these 25 views to generate affordance queries. This sampling strategy ensures comprehensive coverage of the object from multiple perspectives while optimizing computational resources.
For the remaining stages, we utilized all 25 views.

\begin{figure}[ht]
    \centering
    \begin{subfigure}[b]{0.45\linewidth}
        \includegraphics[width=\linewidth]{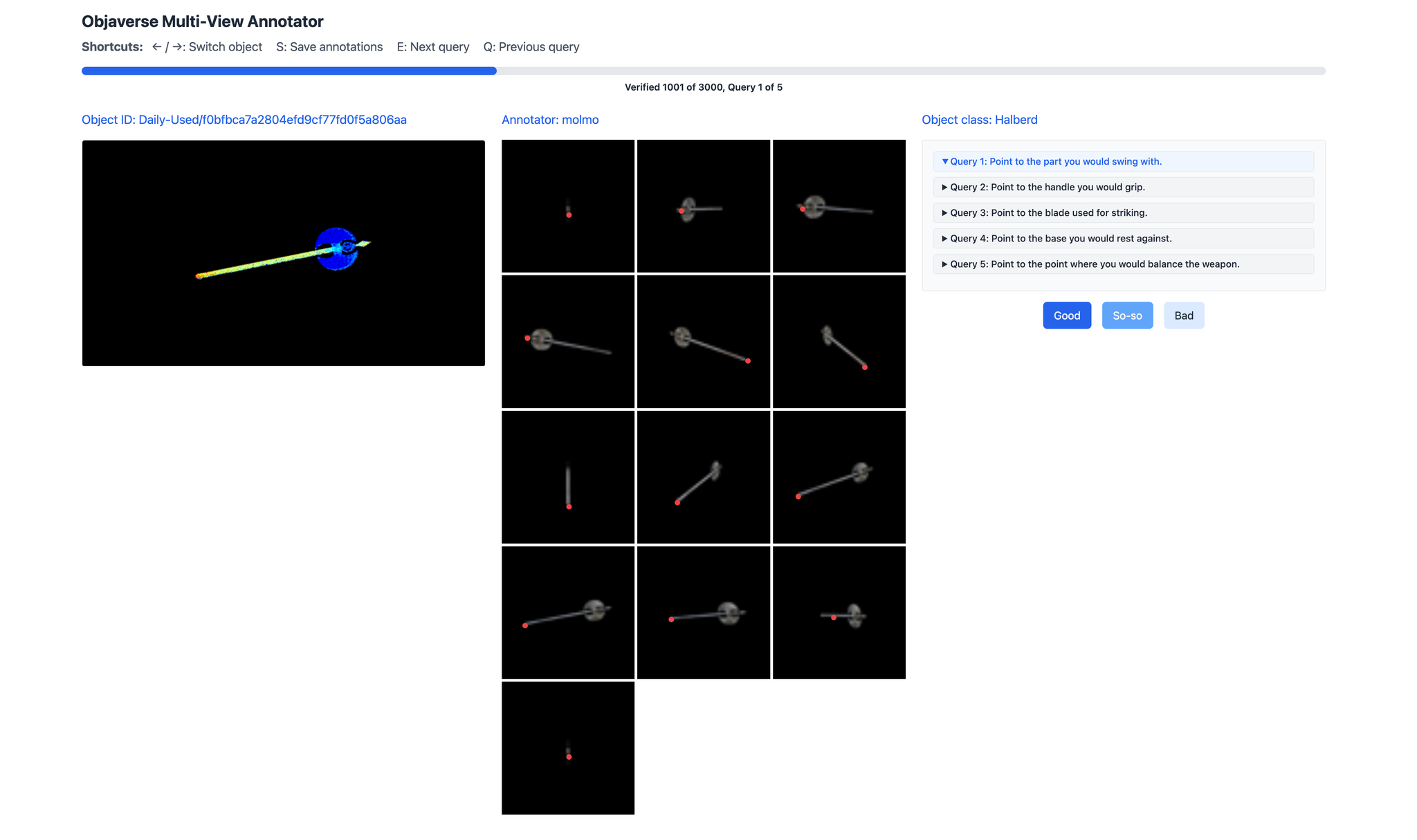}
    \end{subfigure}
    \hfill
    \begin{subfigure}[b]{0.45\linewidth}
        \includegraphics[width=\linewidth]{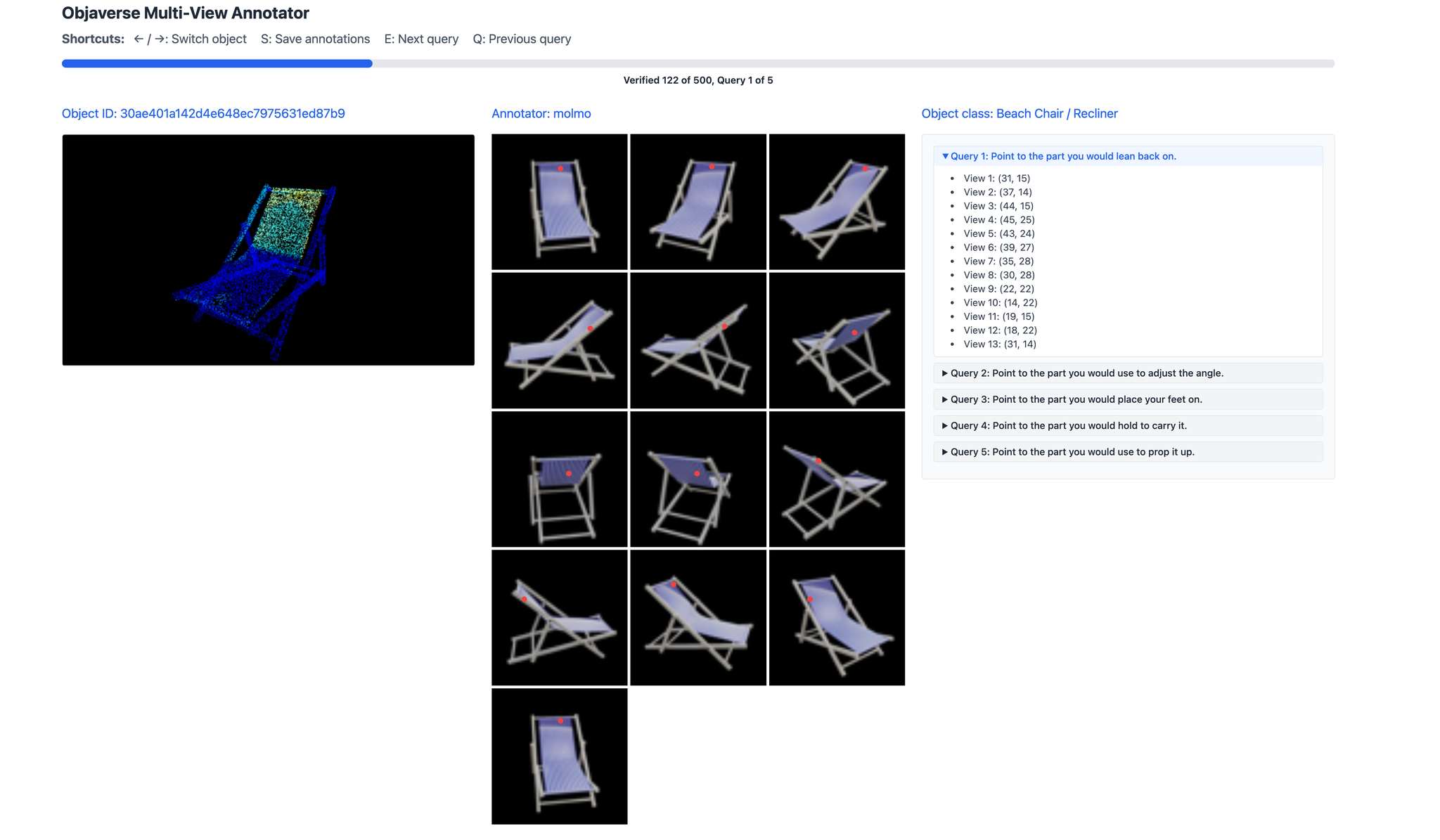}
    \end{subfigure}
    \caption{Web-based interactive viewer for (Left) quality evaluation and (Right) human refinement}
    \label{fig:annotation_ui}
\end{figure}

\noindentbold{Human evaluation}
As discussed in Section 3.3, we instruct human annotators to evaluate the quality of our automatically generated annotation and to refine annotations for those didn't pass the quality check.
We guide the annotators to rate the affordance query-heatmap pairs from three criteria: (1) semantic relevance between the query and object, (2) spatial accuracy of the predicted interaction points, and (3) coverage of the heatmap for the intended affordance.
We provide the annotators with a web-based interactive viewer for screening the affordance query-heatmap pairs and assigning ratings based on the three criteria.
The example of the web-based interactive UI is shown in Figure~\ref{fig:annotation_ui}.

\subsection{3D affordance grounding}

\noindentbold{\modelname-3D architecture}
Our \modelname-3D model leverages the PartField~\cite{liu2025partfield} architecture for processing 3D visual inputs and incorporates Recap-CLIP~\cite{li2024if} for encoding language queries.
Specifically, we adopt a hybrid 3D encoder composed of PVCNN~\cite{liu2019point} and a triplane transformer, as introduced in the official PartField repository.
We used the official pretrained checkpoint and froze the entire vision backbone during training to ensure consistent feature extraction.

For the text branch, we utilize the Recap-CLIP text encoder, which provides enhanced language grounding compared to standard CLIP variants.
The resulting query embeddings are fed into a conditional heatmap decoder that predicts spatial affordance distributions over 3D points.
The decoder augments the 3D point features from the vision encoder with Fourier-based positional encodings and uses them as keys and values in a cross-attention mechanism, where the language embeddings act as queries.
The attended features are then refined through a residual feedforward network (an MLP with skip connections), which outputs the final heatmap over the point cloud.

\noindentbold{Training detail}
To ensure a fair comparison with prior methods, we adopt the same loss formulation as LASO~\cite{li2024laso}.
The model is optimized using a combination of Binary Cross-Entropy (BCE) Loss to handle classification and Dice Loss~\cite{milletari2016v} to improve region-level alignment.
The two losses are summed with equal weights to form the final training objective.
We use the same training setup for both the LASO and \dataname datasets: 50 epochs, batch size of 64, and training on 8 NVIDIA RTX A6000 GPUs.

\subsection{2D affordance grounding}

\begin{table}[t]
    \caption{\small Effect of random background augmentation} 
    \label{tab:augment}

    \begin{subtable}{\linewidth}
    \centering
    \caption{\small Zero-shot evaluation on AGD20K~\cite{agd20k}} 
    \label{tab:agd20k-zeroshotaug}
    \tabcolsep=1.1mm
    \scalebox{0.82}{
        \begin{tabular}{lcc|ccc}
            \toprule
            \multirow{2}{*}{Method} & \dataname & Data & \multirow{2}{*}{KLD~$\downarrow$} & \multirow{2}{*}{SIM~$\uparrow$} & \multirow{2}{*}{NSS~$\uparrow$}  \\
            & pretrain & augmentation & & & \\
            \midrule
            \textit{Seen split} & \\
            \rowcolor{gray!15} \modelname-2D &\checkmark & & 1.493 & 35.5 & 0.920 \\
            \rowcolor{gray!15} \modelname-2D &\checkmark & \checkmark & \textbf{1.426} & \textbf{40.2} & \textbf{0.985} \\ \midrule
            \textit{Unseen split} & \\
            \rowcolor{gray!15} \modelname-2D &\checkmark  & & 1.688 & 31.3 & 0.876 \\
            \rowcolor{gray!15} \modelname-2D &\checkmark  & \checkmark & \textbf{1.571} & \textbf{37.6} & \textbf{1.016} \\
            \bottomrule
        \end{tabular}
    }
    \end{subtable}

    \vspace{3mm} %

    \begin{subtable}{\linewidth}
    \centering
    \caption{\small Fine-tuning on the AGD20K-Full} 
    \label{tab:agd20k-trainaug}
    \tabcolsep=1mm
    \scalebox{0.82}{
        \begin{tabular}{lcc|ccc}
            \toprule
            \multirow{2}{*}{Method} & \dataname & Data & \multirow{2}{*}{KLD$\downarrow$} & \multirow{2}{*}{SIM$\uparrow$} & \multirow{2}{*}{NSS$\uparrow$} \\
            &pretrain & Augmentation & & & \\
            \midrule
            \cellcolor{gray!15} \modelname-2D & \cellcolor{gray!15}\checkmark & \cellcolor{gray!15} & \cellcolor{gray!15} 1.100 & \cellcolor{gray!15} 47.0 & \cellcolor{gray!15} 1.497 \\
            \cellcolor{gray!15} \modelname-2D & \cellcolor{gray!15}\checkmark & \cellcolor{gray!15} \checkmark & \cellcolor{gray!15} \textbf{0.974} & \cellcolor{gray!15} \textbf{51.9} & \cellcolor{gray!15} \textbf{1.645} \\
            \bottomrule
        \end{tabular}
    }
    \end{subtable}

\end{table}

\noindentbold{AGD20K dataset}
    AGD20K-Weak refers to the original AGD20K dataset. The training set consists of 23,083 / 13,323 image-level labels for the Seen / Unseen splits, respectively, while the corresponding test sets contain 1,675 / 540 images. AGD20K-Oneshot refers to the AGD20K dataset for one-shot affordance learning. The training set consists of 50 / 33 images—one per object class—for the Seen / Unseen splits, respectively. The test set is identical to that of AGD20K-Weak. AGD20K-Full is constructed for fully supervised training, following the setup of ~\cite{affordancellm}. The training set consists of 999 images including object classes from the training set of AGD20K’s unseen split, each annotated with dense pixel-level affordance masks. The test set contains 540 images from object classes in the test set of the unseen split. 
    
\noindentbold{Background augmentation in \modelname-2D pretraining stage}
    For the pretraining stage of \modelname-2D, we replace the null background in each rendered image with a randomly selected background from the Background dataset~\cite{gould2009decomposing}. After background replacement, the image is resized to 256×256, randomly cropped to 224×224, and horizontally flipped with a random probability. Table~\ref{tab:augment} summarizes zero-shot performance on AGD20K with and without background augmentation during pre-training, as well as the fine-tuning results on AGD20K-Full.
    Empirically, we observe that background augmentation leads to better generalization compared to pretraining without it. 
    
\noindentbold{\modelname-2D architecture}
    Our \modelname-2D architecture is adapted from OOAL~\cite{li2024one}. Multi-level features from different layers of DINOv2 are aggregated. To focus attention on foreground regions, cross-attention is restricted to the regions indicated by the mask derived from the CLS token. Unlike OOAL, which employs text prompt learning with fixed affordance labels as input, our model takes a single natural language query as the text input without using any text prompt learning.
    
\noindentbold{Training detail}
    We used the CLIP ViT-B/16 as the text encoder and DINOv2 ViT-B/14 as the vision backbone. During pre-training on \dataname, the model is optimized using Adam with a learning rate of 0.001. The training is conducted for 52,000 iterations with a per-GPU batch size of 512 on 7 NVIDIA RTX 3090 GPUs.
    For fine-tuning on AGD20K-Full, we use the Adam optimizer with a learning rate of 0.0001. Training is performed for 400 iterations with a batch size of 512 on a single NVIDIA RTX 3090 GPU. Binary cross-entropy loss is employed consistently in both the pretraining and fine-tuning stages.

\section{Analyses}
\label{sec:analysis}

\subsection{Analyses on Data Annotation Pipeline}
\label{subsec:data_annotation_analysis}

\noindentbold{Effect of data resolution}
\begin{table}[t]
\centering
\caption{\textbf{Performance of \modelname-3D under varying training and test resolutions.} When the training and test resolutions differ, we first perform inference at the training resolution and then interpolate the predicted heatmaps to match the target test resolution.}
\begin{tabular}{l c c c c c}
\toprule
\textbf{Method} & \textbf{Training Res.} & \multicolumn{4}{c}{\textbf{Test Resolution (\# Points)}} \\
\cmidrule(lr){3-6}
 & (\# Points) & \textbf{2048} & \textbf{4096} & \textbf{8192} & \textbf{16384} \\
\midrule
\multirow{2}{*}{\modelname-3D} 
& 2048  & 0.237 & 0.245 & 0.249 & 0.251 \\
& 16384 & 0.265 & 0.278 & 0.283 & 0.287 \\
\bottomrule
\end{tabular}
\label{tab:espresso3d_resolution}
\end{table}

\dataname provides point clouds with a high resolution of 16,384 points, enabling the capture of fine-grained geometric details critical for affordance understanding. This represents a significant improvement over LASO ~\cite{li2024laso}, which provides only 2,048 points resolution. To assess the effect of data resolution, we trained \modelname-3D on \dataname at two resolutions (2,048 vs. 16,384 points) and report the results in \Tref{tab:espresso3d_resolution}. The model trained with high-resolution data consistently outperforms its low-resolution counterpart across all test resolutions, indicating that learning fine-grained geometric details during training is crucial for accurate 3D affordance grounding at any scale.

\noindentbold{Human vs. our annotation pipeline}
\begin{figure}[t!]
    \centering
    \includegraphics[width=0.9\linewidth]{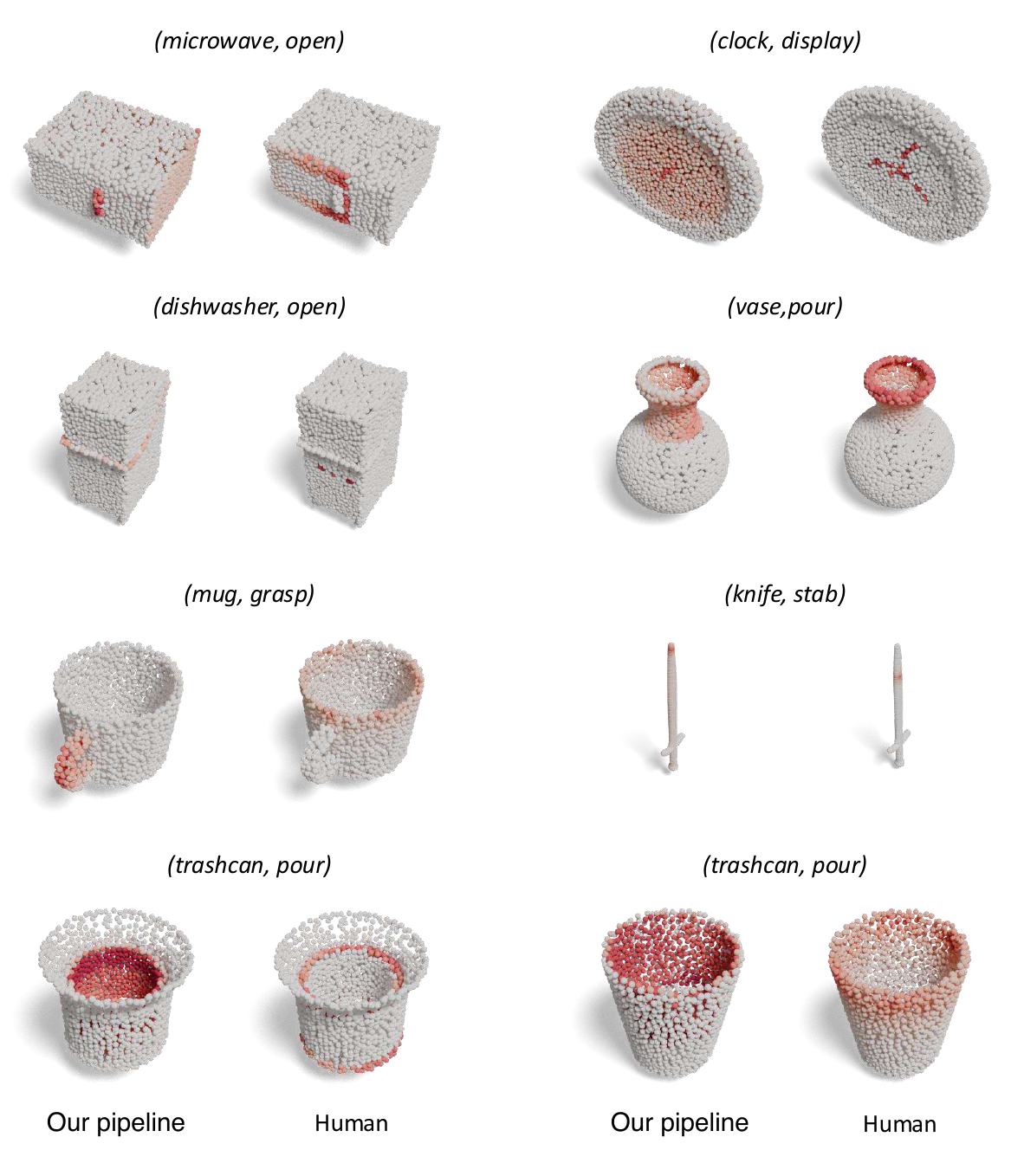}
    \caption{\textbf{Our annotation pipeline vs. Human}. Comparison is conducted on 3D-AffordanceNet~\cite{bahl2023affordances} meshes. First and third columns show affordance heatmaps predicted by~\dataname annotation pipeline, while second and fourth columns are human-annotated.}
    \label{fig:affogato-engine}
\end{figure}

We compare the affordance predictions from our annotation pipeline with human annotations on 3D-AffordanceNet~\cite{bahl2023affordances} meshes. 
Figure~\ref{fig:affogato-engine} presents a qualitative comparison, where the first and third columns display affordance heatmaps generated by our annotation pipeline, while the second and fourth columns show human-annotated ground truth from 3D-AffordanceNet. 
The visual comparison demonstrates that our automated pipeline produces affordance heatmap predictions that closely align with human intuition about object affordances. 
This suggests that our annotation pipeline can serve as a reliable substitute for manual annotation, significantly reducing the time and effort required to create large-scale datasets.

\noindentbold{Failure modes of annotation pipeline}
Despite enabling automatic affordance annotation at scale, our annotation pipeline exhibits several failure modes in its pipeline.
Across all evaluated samples, the dominant failure mode is wrong interaction point prediction in Stage 2, accounting for 34.9\% of all failure cases. In these cases, Molmo struggles with fine-grained grounding, failing to localize the precise functional region and instead pointing to a semantically related but incorrect part of the object, and the error propagates to the subsequent mask generation stage. Stage 1 accounts for another 36.8\% of failure cases in total, comprising incorrect affordance queries (27.6\%) and object misidentification (9.2\%). Despite our use of chain-of-thought prompting, Gemma3~\cite{team2025gemma} occasionally misidentifies the object category or generates affordance queries that are semantically misaligned with the object's actual functionality, and such errors then propagate to later stages. Limited camera coverage accounts for 10.5\% of failure cases across all stages, particularly when the queried object part is occluded or absent from the rendered views. For example, for the query \textit{``Point to the part where you would use to brake the car,''} the brake pedal is often not visible because our multi-view images are captured along a circular trajectory around the object's exterior. Finally, SAM-based mask generation~\cite{mobile_sam,ravi2025sam} contributes 5.2\% of failure cases due to edge bias, which tends to over-emphasize object boundaries and can produce incomplete affordance masks or merge adjacent but functionally distinct parts. We further observe 12.5\% miscellaneous failures.

\subsection{Effect of pretraining-data scale}
\label{subsec:scale}
\begin{table}[t]
\centering
\caption{\textbf{Effect of pretraining-data scale.} \modelname-3D LASO-unseen aIoU and \modelname-2D AGD20K-unseen KLD after pretraining on \dataname subsets of increasing size. Both metrics improve monotonically with scale, while AGD20K-unseen KLD saturates early.}
\begin{tabular}{l cccc}
\toprule
\textbf{Pretrain subset} & \textbf{50K} & \textbf{150K} & \textbf{400K} & \textbf{750K} \\
\midrule
LASO-unseen aIoU\,$\uparrow$  & 18.7 & 19.4 & 19.6 & \textbf{20.8} \\
AGD20K-unseen KLD\,$\downarrow$ & 1.064 & 1.007 & \textbf{0.974} & \textbf{0.974} \\
\bottomrule
\end{tabular}
\label{tab:scale_ablation}
\end{table}

To verify the ``at scale'' claim, we pretrain on \dataname subsets of 50K, 150K, 400K, and 750K object-text pairs and measure downstream generalization (\Tref{tab:scale_ablation}). \modelname-3D LASO-unseen aIoU improves monotonically at every step ($18.7\to19.4\to19.6\to20.8$), and \modelname-2D AGD20K-unseen KLD improves correspondingly while saturating early. The monotonic gains confirm that the improvement is driven by dataset scale rather than by the mere presence of pretraining. We attribute the earlier saturation in the 2D setting to the domain gap between \dataname's synthetic renderings and AGD20K's real-world imagery, which we expect to bound the additional benefit that more synthetic pretraining data can provide.

\subsection{Cross-domain asymmetry}
\label{subsec:asymmetry}
\begin{table}[t]
\centering
\caption{\textbf{Cross-domain asymmetry analysis (\modelname-3D aIoU).} Sub-sampling \textit{Daily-Used} to match the \textit{Furnitures} training volume halves the cross-direction gap ($13.6\to6.1$), identifying training volume as the dominant factor; the residual gap is attributable to test-class coverage.}
\begin{tabular}{l c cc}
\toprule
\textbf{Train} & \textbf{\#obj} & $\rightarrow$\,\textit{Daily} & $\rightarrow$\,\textit{Furn.} \\
\midrule
\textit{Daily-Used} (full) & 121{,}179 & 10.4 & 18.2 \\
\textit{Daily-Used} (sub.) & 8{,}759   & 6.0  & 10.7 \\
\textit{Furnitures}        & 8{,}759   & 4.6  & 20.3 \\
\bottomrule
\end{tabular}
\label{tab:cross_domain}
\end{table}

As shown in the main-paper 3D affordance grounding results (Sec.~5.2, Table~3), the \dataname cross-domain splits exhibit an asymmetry: \modelname-3D trained on \textit{Daily-Used} and evaluated on \textit{Furnitures} reaches 18.2 aIoU, whereas the reverse direction attains only 4.6. We attribute this asymmetry primarily to training volume and secondarily to category coverage. The two splits differ by $\approx$14$\times$ in training volume (121K vs.\ 8.8K objects) and $\approx$4$\times$ in test-class diversity (515 vs.\ 122 classes). As shown in \Tref{tab:cross_domain}, sub-sampling \textit{Daily-Used} to match the \textit{Furnitures} volume halves the cross-direction gap ($13.6\to6.1$), identifying training volume as the dominant factor, since geometry is invariant to sub-sampling. The residual gap reflects class coverage: \textit{Daily-Used}-trained models cover the 122 \textit{Furnitures} classes more readily than the reverse covers 515, so even volume-matched \textit{Daily-Used}\,(sub.) scores higher on \textit{Furnitures} (10.7) than on \textit{Daily-Used} (6.0).

\subsection{Failure cases of \modelname}
\label{subsec:model_analysis}
\begin{figure}[t]
    \centering
    \includegraphics[width=\linewidth]{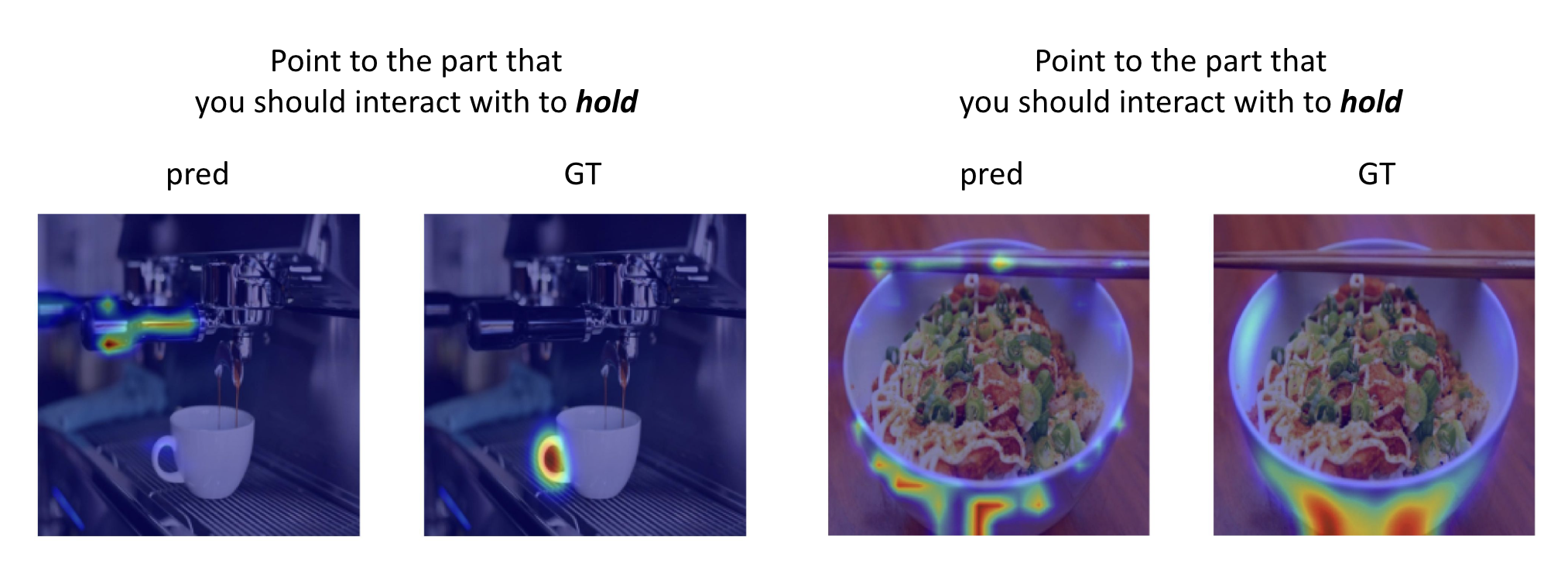}
    \caption{
    Failure cases of \modelname-2D in zero-shot evaluation on AGD20K.
    }
    \label{fig:failure_cases2d}
\end{figure}
 Figure~\ref{fig:failure_cases2d} illustrates failure cases from the zero-shot evaluation of the \modelname-2D model, pretrained on \dataname, on the AGD20K dataset. Images in AGD20K sometimes contain multiple objects, which can result in several plausible regions corresponding to a single affordance. For example, in the left image, the "hold" affordance could refer either to the handle of the coffee machine or to the handle of the cup.
Similarly, in the right image, "hold" could apply to either the chopsticks or the outer surface of the bowl. These cases highlight limitations in the ground truth annotations. To make more precise predictions, it may be necessary to include explicit object information in the prompt, such as “hold the cup” or “hold the bowl.” Our model is capable of accepting arbitrary natural language queries, making it well-suited for resolving such disambiguities effectively.

\section{Limitations}
\label{sec:limitations}
As the \dataname dataset is derived from the Objaverse 3D assets, our data do not contain the background information.
Due to this limitation, we randomly synthesize background on the 2D images as shown in Table~\ref{tab:augment}, which is shown to be helpful when transferred to the real-world images.
Note that our data engine can be extended to the indoor or outdoor scene data to tackle navigation environments, leaving them for future work.

\section{Additional Qualitative Results}
\label{sec:additional_results}

\noindentbold{Qualitative comparison of zero-shot evaluation on AGD20K}
\begin{figure}[t]
    \centering
    \includegraphics[width=\linewidth]{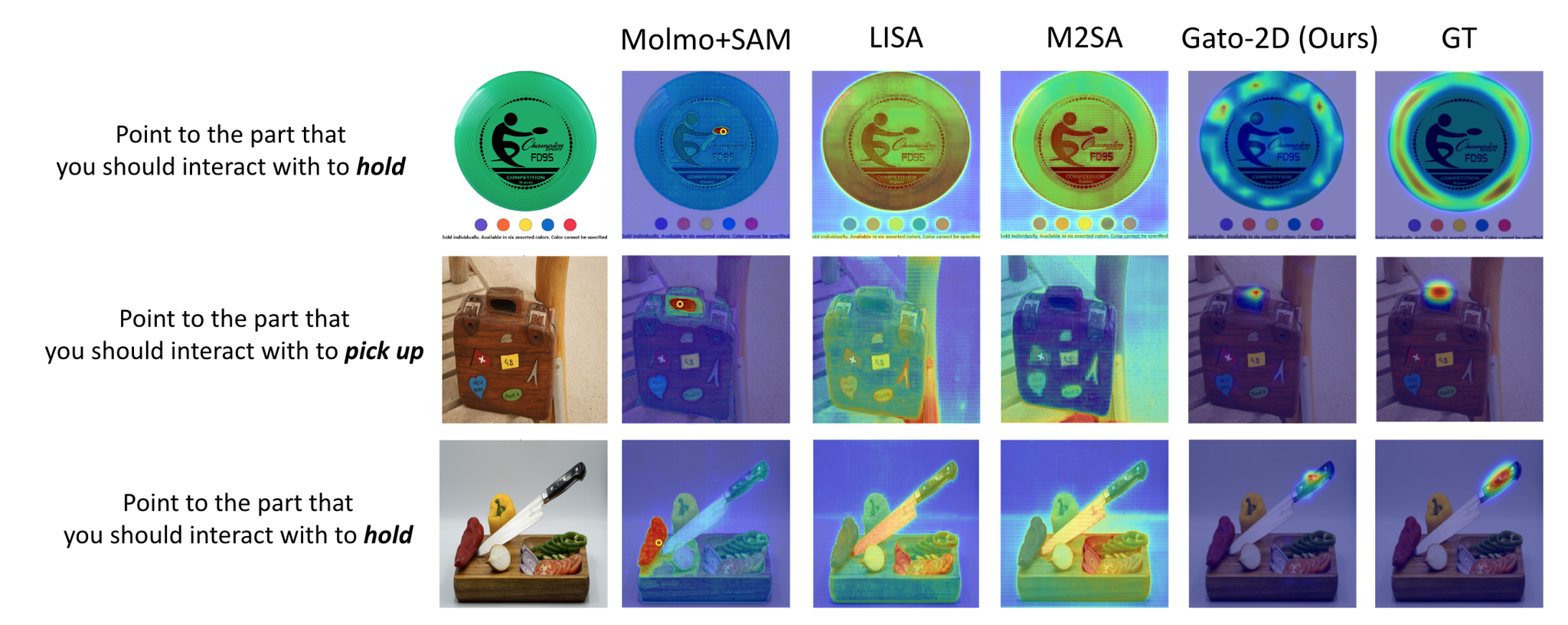}
    \caption{Zero-shot evaluation comparison of \modelname-2D and others on AGD20K~\cite{agd20k}  }
    \label{fig:qual2d_suppl}
\end{figure}

In \Figref{fig:qual2d_suppl}, we visualize qualitative results for 2D zero-shot affordance grounding. It illustrates common failure modes of existing methods~\cite{lisa, jang2025mmr} that capture whole objects rather than precise part-level affordances.
While Molmo+SAM provides rough part-level localization, our model trained on the multi-view aggregated \dataname dataset achieves refined grounding capabilities.

\noindentbold{Stage-wise qualitative results}
\begin{figure}[t]
    \centering
    \includegraphics[width=1.0\linewidth]{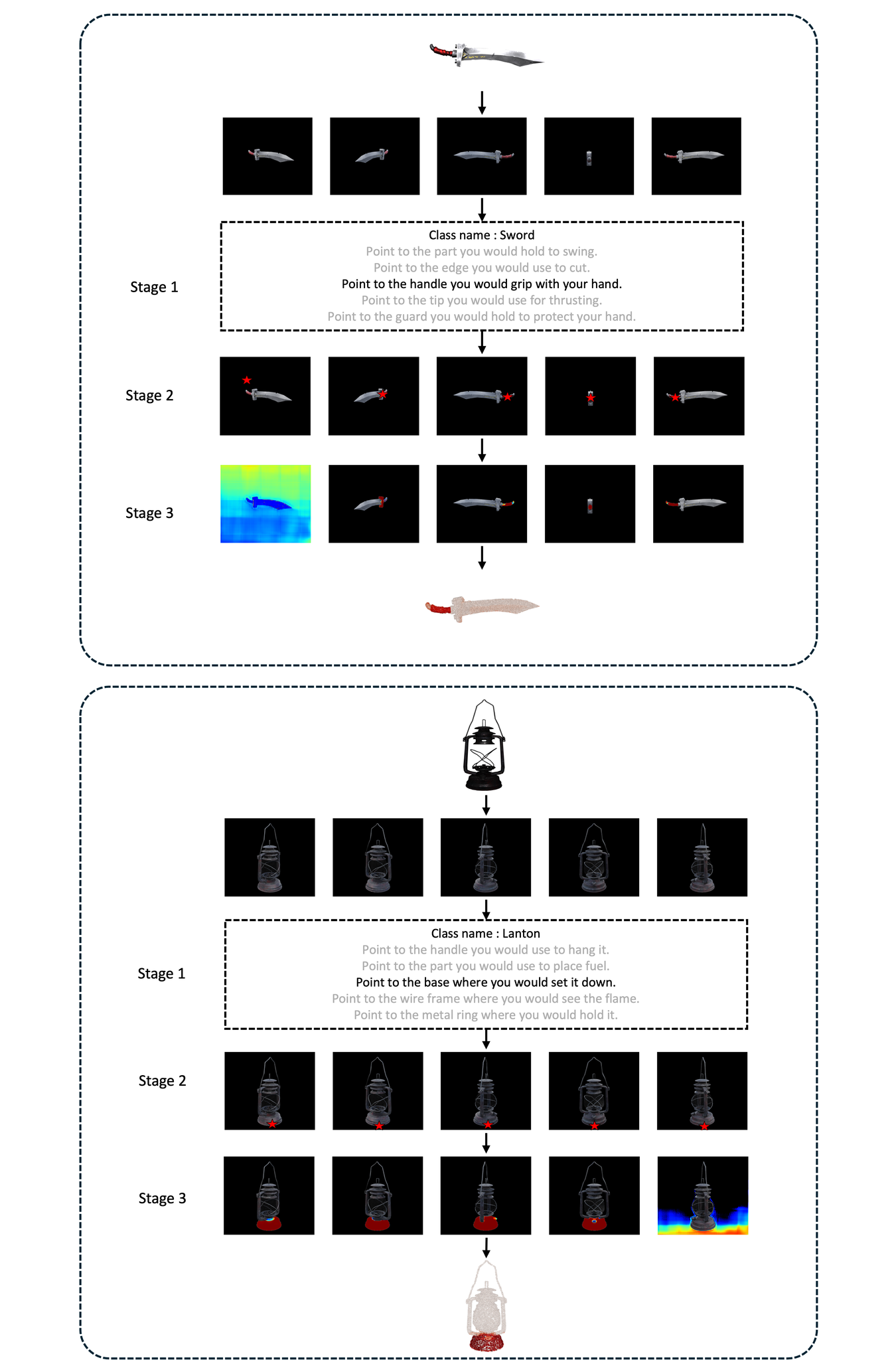}
    \caption{~\textbf{Intermediate qualitative results at each stage of annotation pipeline.} For Stage 2 and Stage 3, we present the results corresponding to the affordance query highlighted in black from Stage 1. In Stages 2 and 3, the visualizations are shown for five representative views, uniformly sampled from the all 25 views.}
    \label{fig:supp_stage_wise}
\end{figure}

\Figref{fig:supp_stage_wise} presents the intermediate qualitative results at each stage of our pipeline.
Stage 1 takes multi-view images as input and predicts the object’s class name together with five candidate queries.
Stage 2 leverages Molmo to point to the affordance region corresponding to each query.
Stage 3 converts Molmo’s pointing locations into pixel-wise masks.
These results highlight the robustness of our pipeline: even when Molmo produces incorrect pointings for some views or MobileSAM generates imprecise masks, the multi-view consensus voting effectively suppresses such errors, yielding an accurate final output.

\end{document}